\documentclass[lettersize,journal]{IEEEtran}
\ifCLASSOPTIONcompsoc
  \usepackage[nocompress]{cite}
\else
  \usepackage{cite}
\fi
\ifCLASSINFOpdf
\else
\fi
\usepackage{amsmath,amsfonts}
\usepackage{algorithmic}
\usepackage{array}
\usepackage{textcomp}
\usepackage{stfloats}
\usepackage{url}
\usepackage{verbatim}
\usepackage{graphicx}
\usepackage{amssymb}
\usepackage{bm}
\usepackage{subfigure}
\usepackage{multirow}
\usepackage{color}
\usepackage{cite}
\usepackage{tabularx}
\usepackage{float}
\usepackage{fdsymbol}
\usepackage[switch]{lineno}
\usepackage{hyperref}
\usepackage{bbding}
\usepackage{booktabs}

\DeclareMathOperator*{\argmin}{arg\,min}

\def\Thetaset{\bm{\Theta}}

\def\N{\mathcal{N}}
\def\NxX{\N_{\x_i,K_1}^{\X}}
\def\NXX{\N_{\X,K_1}^{\X}}

\def\NxXl{\N_{\x_i,K_2}^{\X_l}}

\def\p{\mathbf{p}}
\def\x{\mathbf{x}}
\def\y{\mathbf{y}}
\def\s{\mathbf{s}}
\def\e{\mathbf{e}}

\def\X{\mathbf{X}}
\def\Y{\mathbf{Y}}

\def\E{\mathbf{E}}
\def\S{\mathbf{S}}
\def\R{\mathbb{R}}
\def\V{\mathbf{V}}
\def\D{\mathbf{D}}
\def\W{\mathbf{W}}
\def\L{\mathbf{L}}
\def\u{\mathbf{u}}
\def\t{\mathbf{t}}

\def\G{\mathcal{G}}
\def\b{\mathbf{b}}
\def\M{\mathbb{M}}
\def\A{\mathbb{A}}

\newcommand{\znote}{\textcolor{black}}
\newcommand{\zznote}{\textcolor{black}}

\begin{document}

\title{Perception-Guided Quality Metric of 3D Point Clouds Using Hybrid Strategy}

\markboth{Journal of \LaTeX\ Class Files,~Vol.~14, No.~8, August~2015}%
{Shell \MakeLowercase{\textit{et al.}}: Bare Demo of IEEEtran.cls for Computer Society Journals}
%



\author{Yujie Zhang, Qi Yang, Yiling Xu,~\IEEEmembership{Member,~IEEE}, and Shan Liu,~\IEEEmembership{Fellow,~IEEE}
\thanks{This paper is supported in part by National Natural Science Foundation of China (61971282, U20A20185) and  the Fundamental Research Funds for the Central Universities of China, and STCSM under Grant (22DZ2229005). The corresponding author is Yiling Xu (e-mail: yl.xu@sjtu.edu.cn). }
\thanks{Y. Zhang, Y. Xu are from Cooperative Medianet Innovation Center, Shanghai Jiao Tong University, Shanghai, 200240, China, (e-mail: yujie19981026@sjtu.edu.cn, yl.xu@sjtu.edu.cn)}
\thanks{Q. Yang is with Tencent MediaLab, Shanghai, China (email: chinoyang@tencent.com) }
\thanks{S. Liu is with Tencent MediaLab, Palo Alto, America (email: shanl@tencent.com) }

}

\IEEEtitleabstractindextext{%
\begin{abstract}
Full-reference point cloud quality assessment (FR-PCQA) aims to infer the quality of distorted point clouds with available references. Most of the existing FR-PCQA metrics ignore the fact that the human visual system (HVS) dynamically tackles visual information according to different distortion levels (i.e., distortion detection for high-quality samples and appearance perception for low-quality samples) and measure point cloud quality using unified features.  To bridge the gap, in this paper, we propose a perception-guided hybrid metric (PHM) that adaptively leverages two visual strategies with respect to distortion degree to predict point cloud quality: to measure visible difference in high-quality samples, PHM takes into account the masking effect and employs texture complexity as an effective compensatory factor for absolute difference; on the other hand, PHM leverages spectral graph theory to evaluate appearance degradation in low-quality samples. Variations in geometric signals on graphs and changes in the spectral graph wavelet coefficients are utilized to characterize geometry and texture appearance degradation, respectively. Finally, the results obtained from the two components are combined in a non-linear method to produce an overall quality score of the tested point cloud. The results of the experiment on five independent databases show that PHM achieves state-of-the-art (SOTA) performance and offers significant performance improvement in multiple distortion environments. The code is publicly available at \url{https://github.com/zhangyujie-1998/PHM}.

\end{abstract}

}

\maketitle

\IEEEdisplaynontitleabstractindextext

%
\IEEEpeerreviewmaketitle

\section{Introduction}\label{sec:intro}

\begin{figure}
    \centering
    {\includegraphics[width=1\linewidth]{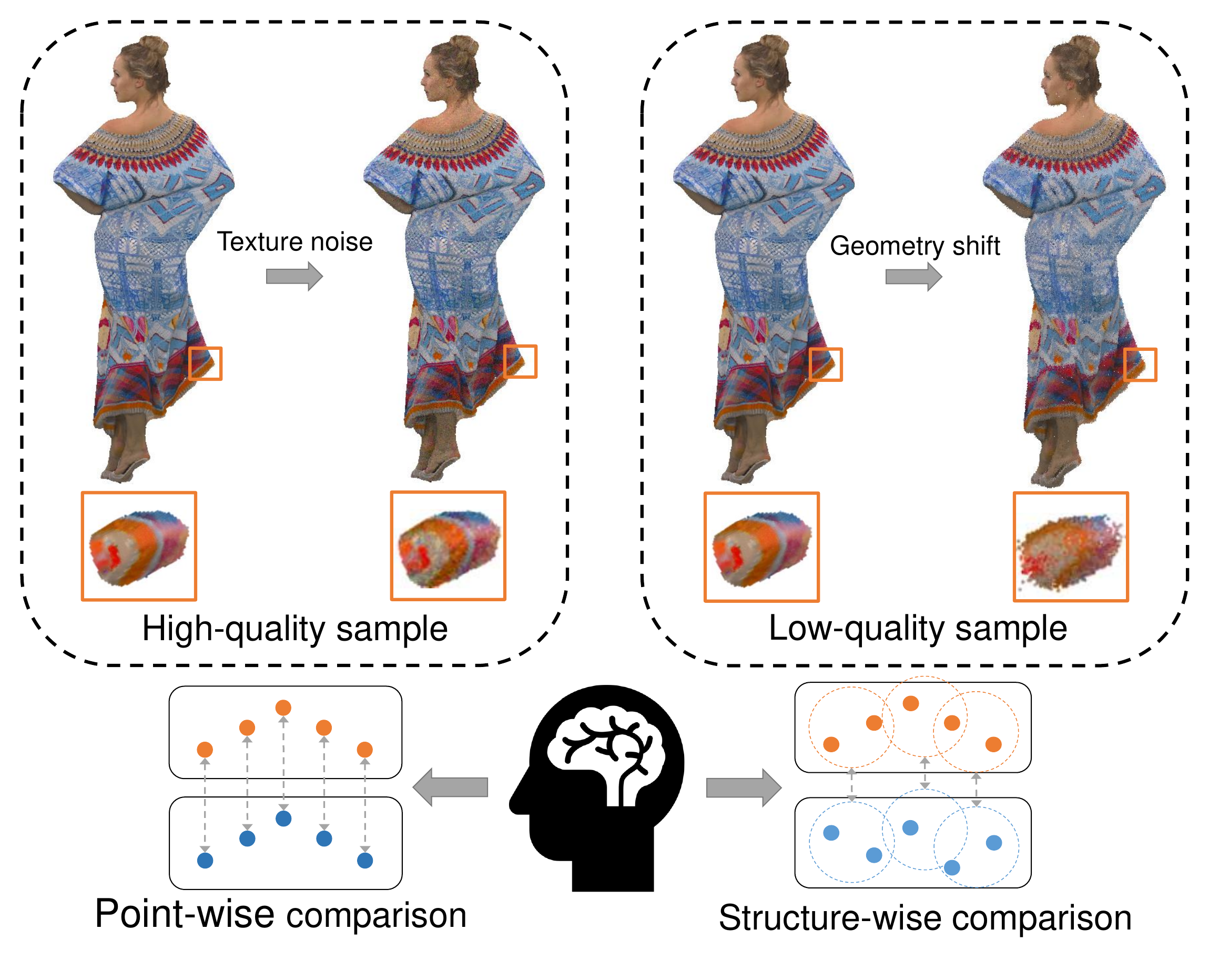}}
    \caption{The core idea of this paper. For the high quality sample in the left box, the HVS tends to use point-wise comparison to detect microscopic differences. For the low quality sample in the right box, the HVS tends to use structure-wise comparison to measure appearance degradation at a large scale.  }
    \label{fig:motivation}
\end{figure}

With recent advances in 3D capturing devices, point clouds have emerged as one of the most prevalent media formats to represent 3D visual content in multiple immersive applications. Point cloud is a collection of points scattered in the 3D space, where each point has spatial coordinates and other additional attributes, such as RGB values and surface normal. Before being delivered to the user client, point clouds are processed through a wide variety of stages, including acquisition, compression, transmission, and rendering. Any stage might cause deterioration in visual quality. To guide the quality of experience (QoE) oriented optimization \cite{liu2021reduced, su2023bitstream,liu2021pqa,lv2024no}, point cloud quality assessment (PCQA) has become one of the most fundamental and challenging problems in both the industry and the academic area. 

PCQA falls into two categories: subjective PCQA by humans and objective PCQA by algorithms designed to correlate with subjective judgment. Although subjective PCQA can provide the correct evaluation, it is usually too time-consuming and expensive to be incorporated into the automatic or real-time processing system. Therefore, it is necessary to develop effective objective PCQA metrics that can replace subjective PCQA, facilitating the application of point clouds. Objective PCQA metrics can be classified into full-reference (FR), reduced-reference (RR) \cite{liu2021reduced,su2023support}, and no-reference (NR) \cite{shan2022gpa,shan2024contrastive, liu2022no,zhu20243dta}, metrics according to the availability of a “perfect quality” reference point cloud. This paper focuses on developing a FR-PCQA metric.

Generally, existing FR-PCQA metrics can be categorized into two classes based on the minimum processing unit: point-wise metrics \cite{MPEGSoft,javaheri2020generalized} and structure-wise metrics \cite{meynet2020pcqm, yang2020inferring, zhang2021ms, alexiou2020towards, diniz2020towards}. Point-wise metrics, such as point-to-point (p2po) and $\rm PSNR_{YUV}$, aim to gauge the absolute difference between the matched points. Given the reference and distorted point cloud, they first match points in one sample with their nearest neighbors in the other sample, and then calculate location/attribute errors over all pairs of points. In comparison, structure-wise metrics assume that structural change in local regions (e.g., gradient \cite{yang2020inferring}, curvature \cite{meynet2020pcqm}) dominates the quality decision, thus quantifying perceptual deformation at a macro level. For this purpose, these metrics initially partition the point cloud into multiple local patches, followed by extracting perceptually significant features. For example, by constructing local graphs around keypoints and considering color values as graph signals, GraphSIM \cite{yang2020inferring} aggregates color gradient moments to estimate visual distortion. In \cite{alexiou2020towards}, pointSSIM utilizes statistical quantities within neighborhoods to characterize the local topology of point clouds.

However, the above metrics, whether they are point-wise or structure-wise, have generally been developed from a holistic perspective regardless of the nature of the distortion, resulting in inconsistent performance among different distortion environments. As shown in Fig. \ref{fig:motivation}, point-wise metrics are more capable of capturing these microscopic distortions (e.g., texture noise) that are usually observed in high-quality samples but usually fail to reflect the macroscopic structural variations (e.g., geometry shift) that cause severe visual degradation. 
This is mainly because the nearest neighbor matching used in point-wise comparison is significantly affected by the geometric fidelity of distorted samples.
In contrast, well-designed structure-wise metrics can effectively seize structural deformations that are observed in low-quality samples, such as alterations in curvature, but usually perform less well over the marginal disturbance. The reason is that distortions at isolated points do not result in substantial disturbances of specific structural features. Overall, the inconsistent performance in various distortion scenarios impairs the generalization of existing metrics, serving as a challenging problem for FR-PCQA.

Several recent studies \cite{schulkin2009cognitive, larson2010most, zhang2015perception}  have shown that the human visual system (HVS) is a highly adaptive system that employs independent processes for various distortion levels. For high-quality samples, distortions are not readily visible and can be tolerated due to the masking effect \cite{hu2015compressed}. Therefore, the HVS seems to employ a detection-based strategy to locate any visible perturbance. For low-quality samples, suprathreshold geometry and texture distortion dominate the quality decision. Therefore, the HVS intends to employ an appearance-based strategy that measures macroscopic structural deformation. On the basis of these insights, it is expected to design a hybrid quality metric to dynamically adopt different visual strategies: for high-quality samples, the hybrid PCQA metric should emphasize the measurement of visible difference by taking into account the masking effect; for low-quality samples, the hybrid PCQA metric should prioritize the evaluation of appearance degradation at a large scale.

In this paper, we propose a novel FR-PCQA metric, called perception-guided hybrid
metric (PHM), which combines the two measurements of visible difference and appearance degradation in an adaptive manner. Specifically, to merge the masking effect into the distortion measurement of high-quality samples, we investigate the relationship between absolute difference and subjective scores, followed by introducing texture complexity as an effective compensation factor. On the other hand, we employ spectral graph theory \cite{chung1997spectral} to evaluate the appearance degradation of low-quality samples. By modeling the spatial graph for local regions, we quantify the geometry appearance degradation by assessing the intrinsic smoothness of the coordinates on graphs. Additionally, from a multi-resolution perspective, we explore the application of spectral graph wavelet transform (SGWT) \cite{hammond2011wavelets} on point clouds, and the weighted co-occurrence matrices (WCM) of graph wavelet coefficients are utilized to evaluate the texture appearance degradation in multiple sub-bands. Finally, an adaptive combination method is used to merge visible difference and appearance degradation, which helps emphasize different components according to various distortion levels. The main contributions are summarized as follows:
\begin{itemize}
    \item Considering the HVS dynamically tackles visual information, we propose a hybrid FR-PCQA metric that combines visible difference and appearance degradation in an adaptive manner.
    \item Given that the masking effect is essential in visual perception, we utilize the texture complexity to modulate the absolute difference in high-quality samples.
    \item By modeling the relationship among points, we assess the appearance degradation via the spectral graph theory for low-quality samples.
    \item Our metric shows reliable performance in five publicly accessible databases. Further analyses reveal the model's robustness across different scenarios.

\end{itemize}
  
The remainder of this paper proceeds as follows. Section \ref{sec:related works} presents the related work. Section \ref{sec:preliminaries} introduces the necessary preliminaries of our metric. The implementation details of the proposed method are presented in Section \ref{sec:proposed method}. Section \ref{sec:experiment} gives the experiment results. Conclusion is drawn in Section \ref{sec:conclusion}.

\section{Related works}\label{sec:related works}

The development of PCQA metrics has emerged as an active research field in recent years. A variety of metrics have been proposed for FR-PCQA, which can be categorized as point-wise metrics and structure-wise metrics based on the minimum processing unit. We give a brief review of these metrics in the following.

\textbf{Point-wise Metrics.}
Early point-wise metrics include point-to-point (p2po), point-to-plane (p2pl) and $\rm PSNR_{YUV}$ \cite{MPEGSoft}. Specifically, p2po and $\rm PSNR_{YUV}$ calculated the errors between coordinates/YUV values of matched points. In comparison, p2pl extended p2po by projecting the error vector along the local normal, which assigned points closer to the surface with smaller errors. Note that these metrics use both the original and distorted point clouds as references and then use the max operation on the resulting quality scores to deliver a final symmetric prediction.

The three aforementioned metrics have been incorporated into MPEG standardization activities because of their computational efficiency. As a consequence, they were widely used and tested by the research community, attracting interest and inspiring submissions to improve their performance. To alleviate the influence of outlier points, Javaheri et al. \cite{javaheri2020generalized} proposed to use the Hausdorf distance for a specific percentage of the data rather than the entire data set.  In \cite{wang2023improving}, Wang et al. incorporated just noticeable difference theory (JND) to derive the noticeable possibility of each point, leading to a better correlation with subjective judgment. 

\textbf{Structure-wise Metrics.}
Structure-wise metrics consider that structural information is critical for visual perception. Early structure-wise metrics intend to extract structural features based on 2D projection. These metrics first project the point cloud onto 2D planes at different viewing angles and then take advantage of the well-developed image processing tools to assist in PCQA. In \cite{alexiou2019exploiting}, Alexiou et al. reported the performance of different 2D structure-wise metrics (e.g., SSIM \cite{wang2004image} and VIFP \cite{sheikh2006image}) by applying them to projected images. In \cite{yang2020predicting},
Yang et al. projected point clouds to obtain texture and depth maps. By utilizing the gradient feature in the depth map, the metric provided better performance than traditional IQA metrics that only used texture maps. 

\begin{figure}
    \centering
    \subfigure[]{\includegraphics[width=\linewidth]{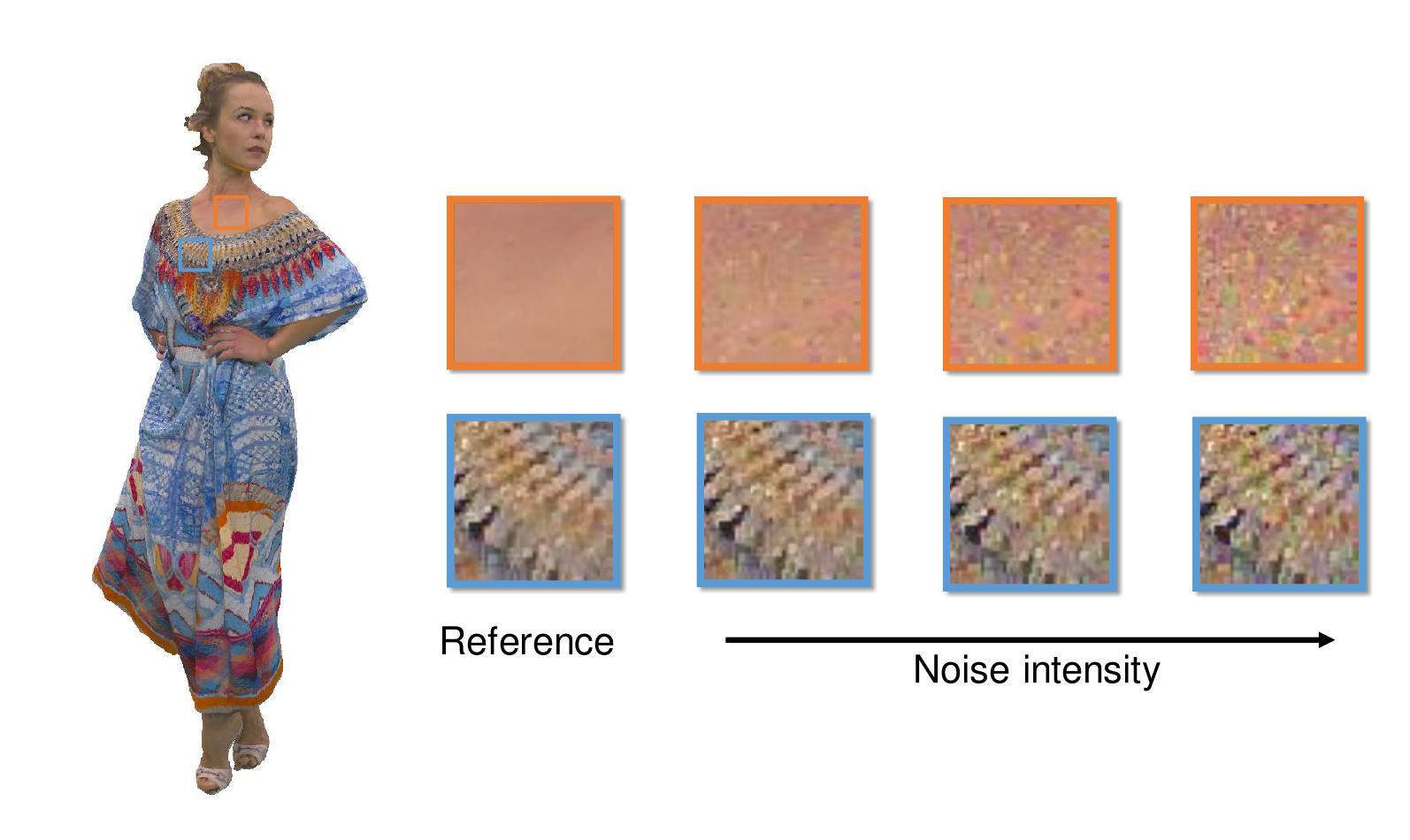}}
    \subfigure[]{\includegraphics[width=\linewidth]{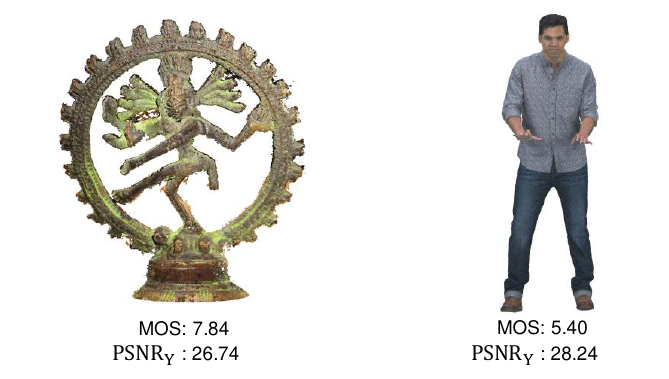}}
    \caption{Two examples of texture masking effect (provided by SJTU-PCQA \cite{yang2020predicting}). (a) Two regions with different texture characteristics impaired by different intensities of texture noise. (b) Two samples impaired by the same intensity of texture noise. }
    \label{fig:texture_masking}
\end{figure}

Projection operation inevitably introduces information loss; therefore, recent FR-PCQA research focuses more on directly extracting 3D structural features.  By employing a linear combination of curvature features and color features extracted from local regions, Meynet et al. \cite{meynet2020pcqm} proposed a point cloud quality metric (PCQM) to assess both geometry and color distortion. In \cite{yang2020inferring}, Yang et al. resorted to graph signal processing theory and measured visual distortions using local graph similarity (GraphSIM). Taking into account three multi-scale characteristics of human perception, i.e., color blurring, detail loss, and size change, a multi-scale version of GraphSIM was proposed by Zhang et al.  \cite{zhang2021ms}. In \cite{diniz2020towards}, Diniz et al. extended the classical 2D local binary pattern (LBP) to voxelized point clouds. PointSSIM \cite{alexiou2020towards} used statistical quantities within the neighborhoods to characterize the local appearance of point clouds.

{\bf Summary.}
As suggested in Section \ref{sec:intro}, most existing metrics are holistically designed, overlooking the fact that the HVS dynamically processes visual signals in response to different distortion levels. Therefore, to achieve an improved correlation with subjective perception, we propose a hybrid FR-PCQA metrics to simulate the adaptive mechanism of the HVS.

\section{Preliminaries}\label{sec:preliminaries}
In this section, we elucidate some basic concepts. We first present the formulation of the FR-PCQA problem. Subsequently, we detail the concept of texture masking effect and spectral graph theory that would be considered in our hybrid metric.

\subsection{Problem Formulation of FR-PCQA}\label{sec:problem_formulation}
Let $\X$ be a 3D point cloud with $N$ points: $\X = \{\x_1, \cdots, \x_N\} \in \R^{N\times6}$, where each $\x_i \in \R^6$ is a vector with 3D coordinates and three-channel color attributes, therefore, $\x_i=[X, Y, Z, R, G, B]\equiv[\x_i^O, \x_i^I]$, where $\x_i^O=[X, Y, Z]$ and $\x_i^I=[R, G, B]$. The superscript “O” stands for geometric \textit{occupancy}, and “I” stands for color \textit{intensity}.

\begin{figure}
    \centering
    \subfigure[]{\includegraphics[width=0.9\linewidth]{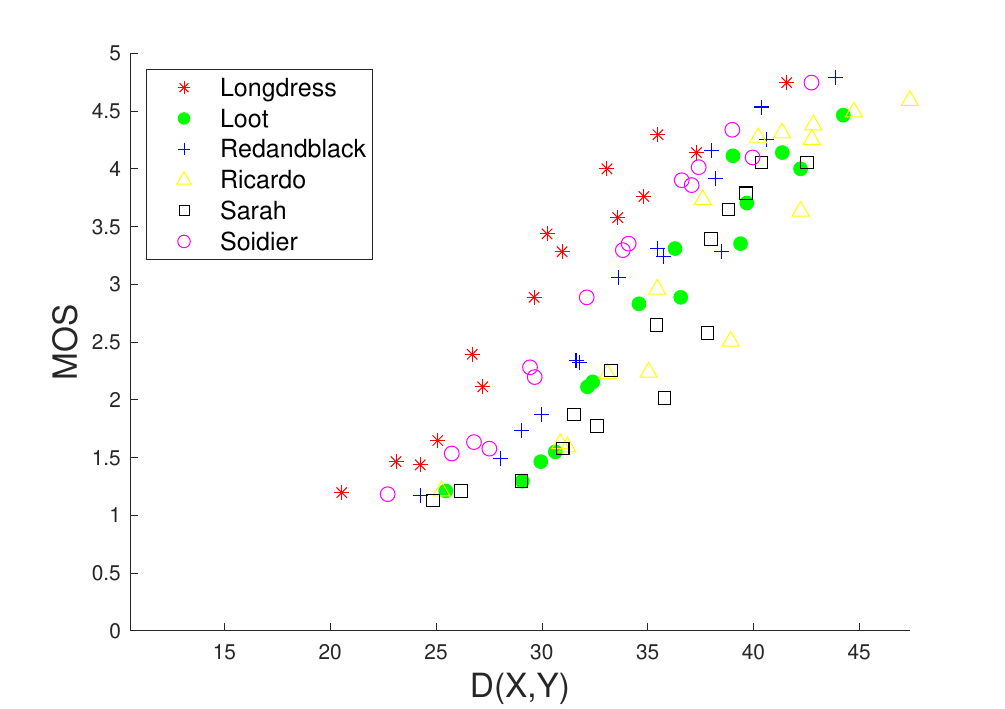}}
    \subfigure[]{\includegraphics[width=0.9\linewidth]{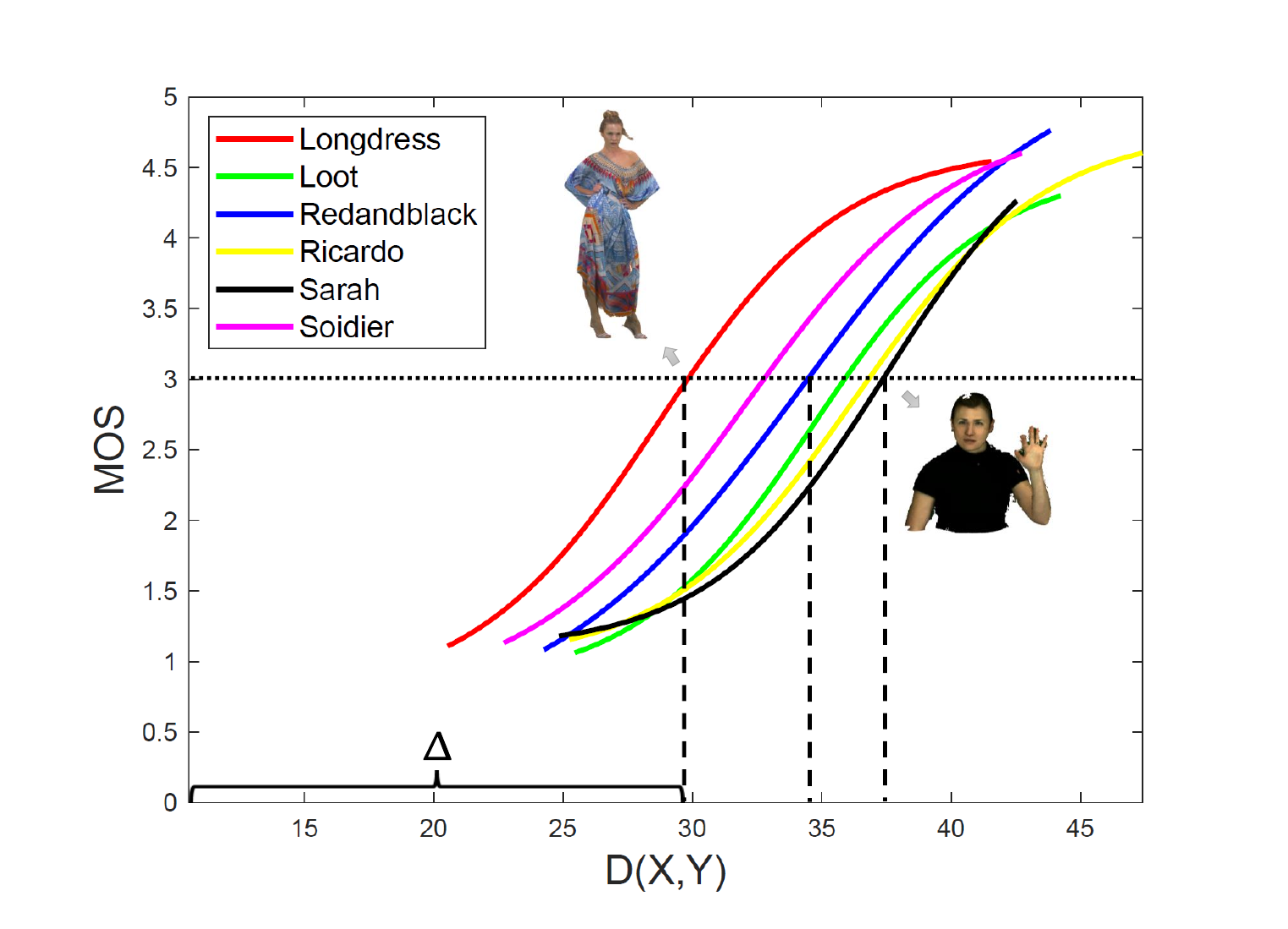}}
    \caption{Investigation of the texture masking effect in FR-PCQA. (a) Scatter plot of MOS vs. $\rm PSNR_{Y}$. (b) Plot of fitted curves and each curve corresponds to one reference point cloud.  }
    \label{fig:texture_masking_plot}
\end{figure}

Given a reference point cloud $\X$ and its distorted version $\Y$,  FR-PCQA intends to evaluate the visual quality of the distorted $\Y$ with respect to its reference $\X$. Mathematically, the purpose of FR-PCQA is to find a measure $D(\cdot)$ that satisfies

\begin{equation}\label{eq:hv_formulation}
  \psi(D(\X,\Y))=q,
\end{equation}
where $q$ denotes the subjective quality score called mean opinion score (MOS) acquired from subjective experiments; $\psi(\cdot)$ denotes the non-linear transform that maps the objective value $D(\X,\Y)$ into the same scales as MOS.

\subsection{Texture Masking Effect in FR-PCQA}\label{sec:texture_masking}
The texture masking effect denotes the reduction in visibility of a certain distortion due to the presence of complex textures. Fig. \ref{fig:texture_masking} shows two examples of the texture masking effect.  Fig. \ref{fig:texture_masking} (a) shows a point cloud impaired by different intensities of texture noise. \znote{We can see that, regardless of the noise intensity, distortions on complex texture regions (e.g., blue box) are generally harder to perceive than that on homogeneous regions (e.g., orange box).} Moreover, Fig. \ref{fig:texture_masking} (b) depicts two samples disturbed by the same intensity of color noise,
where their MOS and luminance-base PSNR (denoted as $\rm PSNR_Y$) are also provided. Similar to $\rm PSNR_{YUV}$, the $\rm PSNR_Y$ can be calculated via a symmetric operation, that is 
\begin{equation}\nonumber
\begin{aligned}
    \mathrm{PSNR_Y}  
   &= 10 \log_{10}(\frac{255^2}{\mathrm{max}(d_{\X2\Y},d_{\Y2\X})}),\\
   d_{\X2\Y} &={\rm mean}(\sum\nolimits_{\x_i\sim \y_j}[(\x_i^I)_{Y}-(\y_j^I)_{\rm Y}]^2)),
\end{aligned}
\end{equation}
where $d_{\X2\Y}$ is a asymmetric measurement, and $\x_i\sim \y_j$ indicates $\y_j$ is the nearest neighbor of $\x_i$ in $\Y$. We can see that the two point cloud 
share similar $\rm PSNR_Y$ values under the same intensity of noise. However, the left point cloud, characterized by extensive texture regions, exhibits a higher MOS in comparison to the right point cloud with abundant flat regions. The above instances illustrate the texture complexity (reflecting the amount of texture information) of point clouds significantly influences the quality prediction of its distorted version. Therefore, defining the texture complexity of the reference $\X$ as $C(\X)$, the paradigm in Eq. \eqref{eq:hv_formulation} can be re-formulated as 
\begin{equation}\label{eq:texture_masking_1}
    \psi(\phi(D(\X,\Y),C(\X)))=q,
\end{equation}
where $\phi(\cdot)$ represents one compensation function that can modulate the original $D(\X,\Y)$ with $C(\X)$ to achieve better perceptual correlation. Note that the new paradigm in Eq. \eqref{eq:texture_masking_1} is mostly effective in measuring subtle visual difference, since severe distortions typically degrade the texture structure of samples, making the texture masking effect negligible \cite{ma2013reduced}.

Investigating how $C(\X)$ adjusts a particular $D(\X,\Y)$ (that is, finding $\phi(\cdot)$) is critical for incorporating the texture masking effect in FR-PCQA. In the paper, we set $D(\X,\Y)$ as $\rm PSNR_Y$ because it is independent of the intrinsic characteristics of the references. To make a concrete analysis, the scatter plot of $\rm PSNR_Y$ versus MOS on ICIP2020 database \cite{perry2020quality} is illustrated in Fig. \ref{fig:texture_masking_plot} (a), where the same type of marks denotes the distorted sample group sharing a particular reference point cloud. We can see that each group of distorted samples has a similar MOS versus $\rm PSNR_Y$ relation. Then, using a non-linear logistic transform $\psi(\cdot)$ \cite{video2003final}, we illustrate the fitted curves of each distorted sample group in Fig. \ref{fig:texture_masking_plot} (b). It is observed that each fitted curve shares a similar shape, but different horizontal displacements, denoted by $\Delta$, that is, $\phi(D(\X,\Y),C(\X))=D(\X,\Y)-\Delta$. Meanwhile, we notice that the curves corresponding to references with high texture complexity (e.g., the "Longdress" reference) have smaller horizontal displacements than those with low texture complexity (e.g., the "Sarah" reference). Therefore, a reasonable assumption is that the horizontal displacement of each curve is determined by the texture complexity of each reference, i.e., $\Delta=f(C(\X))$. Finally, we can rewrite Eq. \eqref{eq:texture_masking_1} as follows
\begin{equation}\label{eq:texture_masking_2}
      \psi(D(\X,\Y)-f(C(\X)))=q.
\end{equation}
Eq. \eqref{eq:texture_masking_2} provides a solution to incorporate the texture masking effect into FR-PCQA. Thus, our work intends to exploit its utilization in assessing distortions of high-quality samples.

\subsection{Spectral Graph Theory and Wavelets on Graphs}\label{sec:gsp_theory}

Spectral graph theory \cite{chung1997spectral} provides generalized concepts and tools to analyze irregular data, such as \textit{graph Fourier transform} (GFT), \textit{inverse graph Fourier transform} (IGFT) and \textit{spectral graph wavelet transform} (SGWT). Due to the non-Euclidean structure of point cloud data, traditional image processing techniques are not applicable. Therefore, spectral graph theory plays an increasingly important role in point cloud processing \cite{hu2021graph}.

\textbf{Graph and Graph Signal}.
A graph $\G=(\mathcal{V},\mathcal{E})$ is composed of a vertex set $\mathcal{V}$ and a edge set $\mathcal{E}$. Generally, for a graph constructed from a point cloud $\X$ with $N$ points, the vertices refer to 3D points in the point cloud. Furthermore, the weighted edges between pairs of vertex are encoded by local geometric information to generate the adjacency matrix $\W\in \mathbb{R}^{N\times N}$, where $w_{i,j}$ is the weight assigned to the edge between $\x_i$ and $\x_j$. From $\W$, we can define the Laplacian matrix $\L$  as $\L=\D-\W$,The degree matrix $\D$ is a diagonal matrix with $d_{i,i}=\sum_jw_{i,j}$.

For a graph $\G$, a \textit{graph signal} $\u\in\R^n$ is defined as $\u=[u_1;u_2;\cdots;u_n]^T$,
where $u_i$ denotes the signal value on the $i$-th vertex.

\textbf{Graph Fourier Transform}.
A Fourier transform performs the expansion of a
signal into a Fourier basis of signals. In the spectral graph theory, the Fourier basis is defined by eigen-decomposition of the graph Laplacian matrix $\L$, i.e., $\L=\V\bm{\Lambda}\V^{T}$,
where the eigenvector $\V$ is called \textit{graph Fourier basis}, and the eigenvalue matrix $\bm{\Lambda}\in \mathbb{R}^{N\times N}$ is a diagonal matrix of corresponding eigenvalues $\lambda_1,\ldots,\lambda_N$ of $\L$ that  are called \textit{frequencies} on the graph. Note that the eigenvalues are arranged in ascending order (i.e., $\lambda_1\leq\lambda_2\leq\ldots\leq\lambda_n$). We set $\lambda_1$ as the lowest frequency and $\lambda_n$ as the highest frequency.

Based on the graph Fourier basis $\V$, the graph Fourier transform of the graph signal $\u$ is
\begin{equation}\nonumber
	\hat{\u}=\mathrm{GFT}(\u)=\V^{T}\u.
\end{equation}
The vector $\hat{\u}=[\hat{u}_1;\hat{u}_2;\cdots;\hat{u}_n]\in\R_n$ describes the spectral response of the graph signal $\u$. Furthermore, the inverse graph Fourier transform (IGFT) of $\hat{\u}$ can be formulated as
\begin{equation}\nonumber
	{\u}=\mathrm{IGFT}(\hat{\u})=\V\hat{\u}.
\end{equation}

\begin{figure*}
    \centering
    \includegraphics[width=0.95\linewidth]{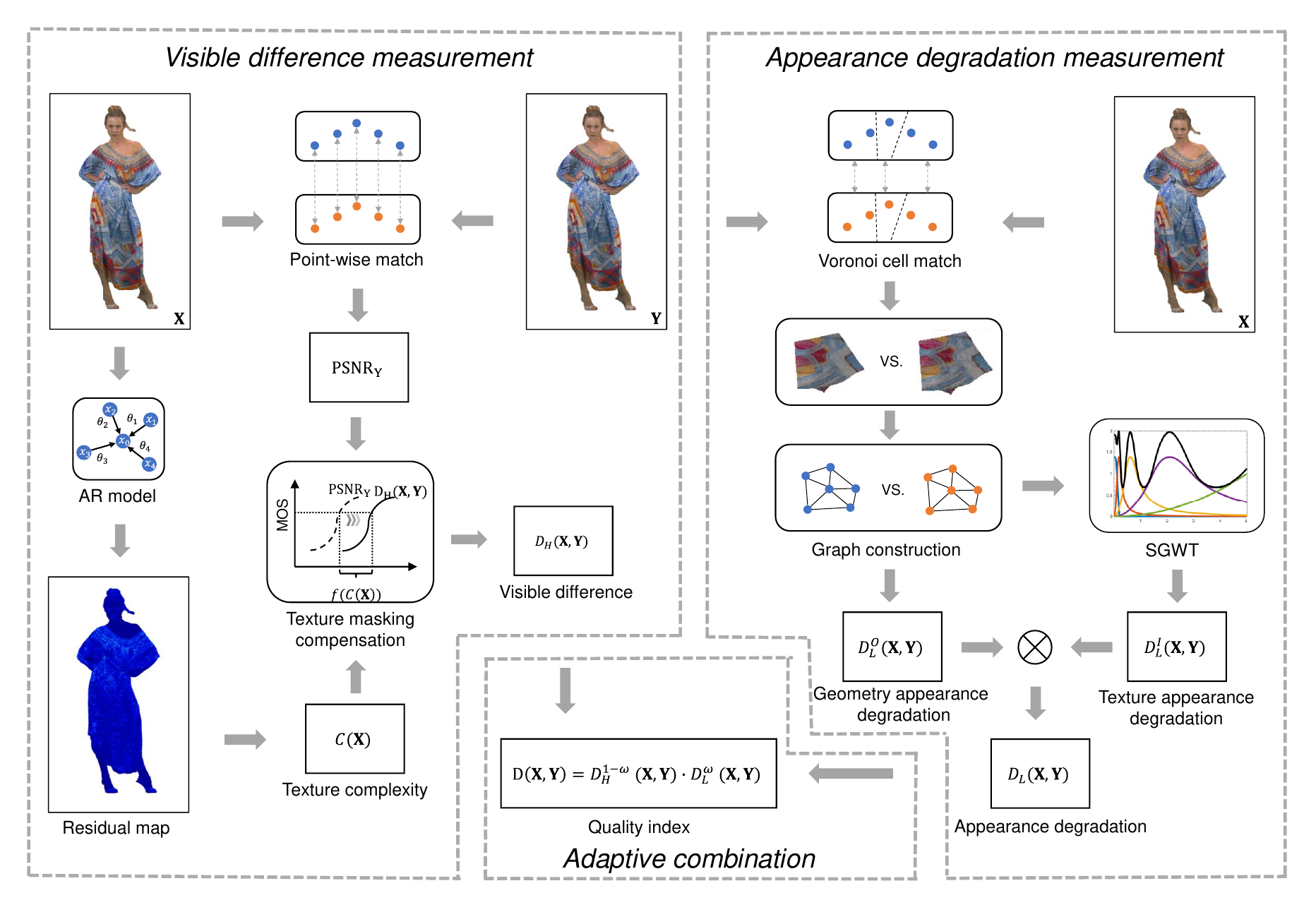}

    \caption{Flowchart illustration of the proposed hybrid quality metric for point clouds, where $\X$ and $\Y$ denote the reference and distorted point clouds, respectively. To simulate the dynamic strategy adpoted by the HVS, the metric contains three parts: visible difference measurement, appearance degradation measurement, and adaptive combination.}
    \label{fig:framework}
\end{figure*}

{\bf Spectral Graph Wavelet Transform.}
GFT is not localized in space, since each eigenvector of $\u$ is on the entire spatial domain. Therefore, Hammond et al. \cite{hammond2011wavelets} proposed the SGWT to address this issue by decomposing $\u$ into a linear combination of basis functions that are localized both in space and frequency. Specifically, SGWT is determined by two operators: wavelet function $g(\cdot)$ and  scaling function $h(\cdot)$. 
The wavelet function $g(\cdot)$ serves as a series of band-pass filters $\{g(t_c\lambda)\}_{c=1}^C$ controlled by one scale parameter set  $\t=\{t_c\}_{c=1}^C$  to decompose $\u$ into multiple sub-bands, analogous to the mother wavelet in traditional wavelet analysis. Mathematically, the wavelet defined by $g(\cdot)$ can be formulated as
\begin{equation}\label{eq:wavelet_function}
\u_c=\V g(t_c\mathbf{\Lambda})\V^T\u=\mathrm{IGFT}(g(t_c\mathbf{\Lambda})\mathrm{GFT}({\u})),
\end{equation}
where $\u_c$ denotes $c$-th wavelet sub-band coefficient of $\u$ with the scale parameter $t_c$, and $g(t_c\mathbf{\Lambda)}$ represents a diagonal matrix whose $i$-th diagonal element is $g(t_c\lambda_i)$. The scale parameter set $\t$  controls the localization of the transform.

The scaling function $h(\cdot)$ is used to stably represent the low-frequency content of $\u$, analogous to the father wavelet in classical wavelet analysis. Mathematically, the wavelet defined by $h(\cdot)$ can be formulated as
\begin{equation}\label{eq:scaling_function}
    \u_0=\V h(\mathbf{\Lambda})\V^T\u=\mathrm{IGFT}(h(\mathbf{\Lambda})\mathrm{GFT}({\u})),
\end{equation}
where $\u_0$ denotes the low-pass wavelet coefficient of $\u$.

The spectral graph theory has emerged as an effective tool in point cloud processing due to its capability of tackling irregular data, which contributes to seizing structural information in FR-PCQA. Furthermore, the SGWT holds significant potential because it can provide a multi-resolution perspective that closely aligns with the processing mechanisms of the primary visual cortex \cite{field1999wavelets}. Therefore, our work aims to utilize spectral graph tools, particularly the SGWT, to model the appearance degradation of low-quality samples.

\section{Proposed Method}\label{sec:proposed method}
In this section, we first overview the proposed method in Section \ref{sec:overview}. Then, we give the detailed implementation for each parts.

\subsection{Overview}\label{sec:overview}

The aforementioned analysis reveals that perceptual quality sensation is complex, and the HVS tends to adopt different strategies for various distortion levels. To address the problem, we propose a hybrid FR-PCQA metric that adaptively tackles visual information, which is composed of three parts:
\begin{itemize}
    \item {\bf Visible Difference Measurement:} The HVS tends to detect these localized perturbances in high-quality samples, where the texture masking effect significantly influences the visibility of distortions. Therefore, we first derive the texture complexity of the reference point clouds based on an auto-regressive (AR) model. Next, we investigate the relationship between the horizontal displacements of the MOS-$\rm PSNR_Y$ curves presented in Section \ref{sec:texture_masking} and the texture complexity. Finally, an effective masking compensation strategy is  employed to modulate $\rm PSNR_Y$.
    \item {\bf Appearance Degradation Measurement:} Distortions in low-quality samples often have a significant impact on the perception of their appearance at a macroscopic level, making them more susceptible to well-designed structure-wise metrics. To exploit the structural relationship among points, we first divide the point clouds into multiple non-overlapping patches and model the graph structure for each patch. We then extract the variation of geometry signals on graphs to characterize the geometry appearance degradation. Furthermore, we measure texture appearance degradation via weighted co-occurrence matrices of spectral graph wavelet coefficients. The index of appearance degradation is obtained by multiplying the two terms.
    \item {\bf Adaptive Combination:} Given that distorted samples can contain a mixture of high-quality and low-quality regions, the final objective quality index is derived by an adaptive combination of the above two measurements.
\end{itemize}

Fig. \ref{fig:framework} presents the overall workflow of the proposed method. The workflow takes the distorted point cloud $\Y\in\R^{M\times 6}$ and its reference $\X\in\R^{N\times 6}$ as input. After calculating the texture complexity of $\X$ (i.e., $C(\X)$), the visible difference can be described as
\begin{equation}
    D_H(\X,\Y) = {\rm PSNR_Y} - f(C(\X)).
\end{equation}
To quantify the appearance degradation, the geometry and texture deformation components are calculated separately, noted as $D_L^O(\X,\Y)$ and $D_L^I(\X,\Y)$. The appearance degradation is described as
\begin{equation}\label{eq:apperance_fusion}
    D_L(\X,\Y) = \sqrt{D_L^O(\X,\Y)\cdot D_L^I(\X,\Y)},
\end{equation}
\znote{where we follow \cite{wang2004image} fuse the two components via multiplication and utilize $\sqrt{\cdot}$ to keep the power at 1.}

Finally, the workflow follows \cite{larson2010most} to fuse the above two terms as 
\begin{equation}\label{eq:hybrid_model}
    D(\X,\Y)=D_H^{1-\omega}(\X,\Y)\cdot D_L^{\omega}(\X,\Y),
\end{equation}
where $\omega$ is expected to flexibly adjust the two components to adapt to human perception.

\subsection{Visible Difference Measurement}

\subsubsection{Texture Complexity Calculation}
The texture complexity of a point cloud is highly related to its self-description capability \cite{zhang2023tcdm}. Specifically, points in homogeneous regions tend to have a stronger correlation with neighbors than those in texture regions. Therefore, we can follow this idea to measure the texture complexity.

Explicitly, we employ the auto-regressive (AR) model to assess the self-descriptive capability of point clouds, given its favorable local adaptability and low computational cost. For one point $\x_i$ in $\X$, the AR model leverages its neighbors to make a self-description for itself, i.e.,
\begin{equation}\label{eq:VAR}
    f(\x_i) = \sum\nolimits_{\x_j\in\NxX}\theta_{j}f(\x_j)+\e_i=\Thetaset f(\NxX)+\e_i,
\end{equation}
where $f(\x_i)\in\R^d$ represents the feature of $\x_i$, accounting for the luminance value $(\x_i^I)_{L}$ in Eq. \eqref{eq:VAR}.  $\NxX \subset\X$ denotes $\x_i$'s $K_1$ closest points in $\X$, where $K_1=20$ is empirically determined. $f(\NxX)=[f(\x_1);f(\x_2);\cdots;f(\x_{K_1})]\in\R^{K_1}$ denotes the feature matrix of the neighbors. $\Thetaset=[\theta_1,\theta_2,\cdots,\theta_{K_1}]\in\R^{K_1}$ denotes the parameter set of AR model and $\e_i$ represents the residual error.

\begin{figure}
    \centering
    \includegraphics[width=\linewidth]{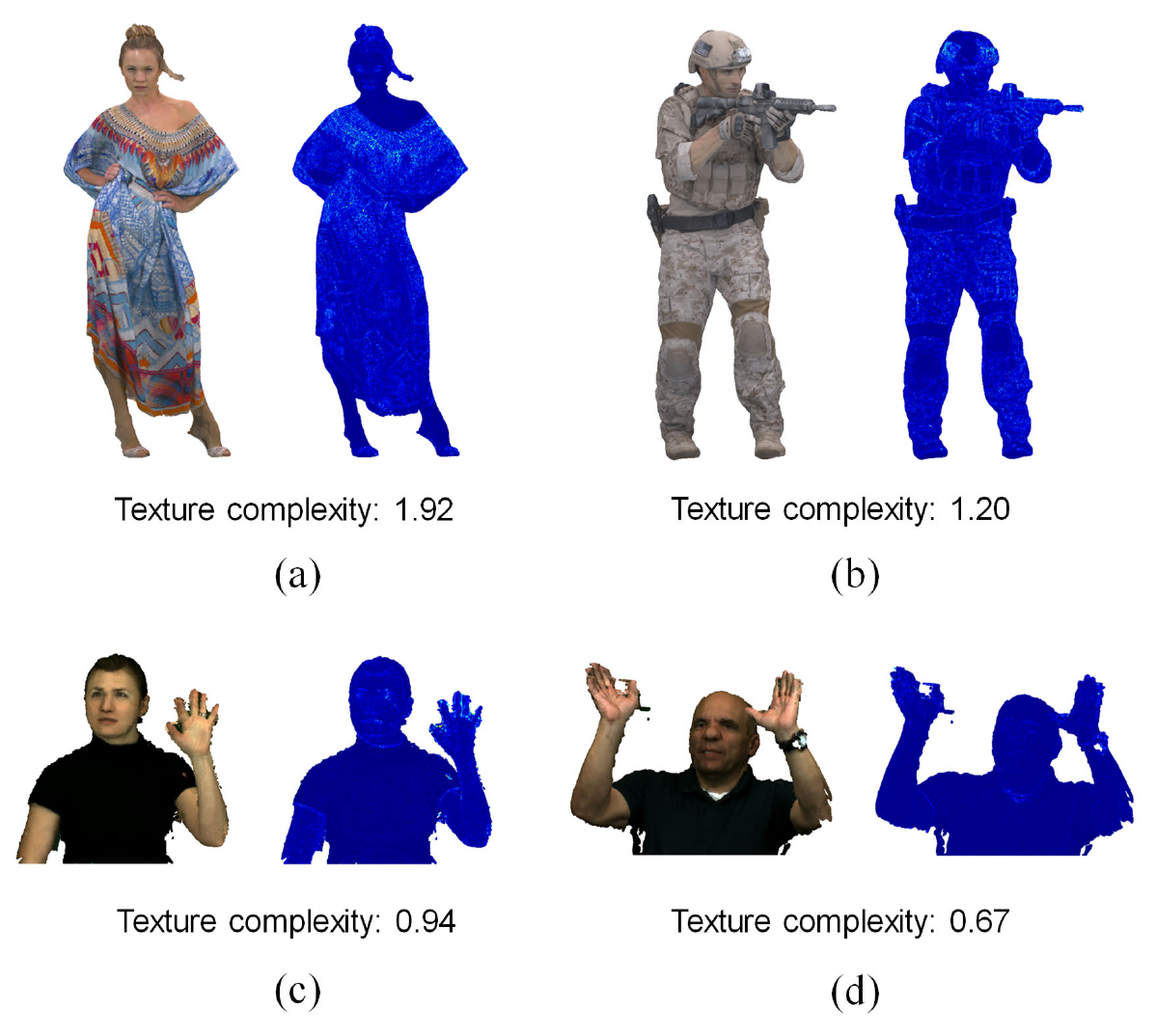}

    \caption{Exemplified examples of the proposed texture complexity evaluation: (a) "Longdress"; (b) "Soldier"; (c) "Sarah"; (d) "Ricardo". The residual error map and the texture complexity value are provided for each sample. Best viewed when zoomed in. }
    \label{fig:AR_residual}
\end{figure}

Applying the above AR model for all points in $\X$, we have
\begin{equation}\label{eq:SA-VAR matrix}
    f(\X)=\Thetaset f(\NXX)+\E,
\end{equation}
where $f(\X)$, $\E \in\R^{N}$ denote the feature and noise matrix of $\X$, and $f(\NXX)\in\R^{K_1 \times N}$ denotes the neighbor feature matrix of  $\X$. In order to determine the optimal AR parameter $\tilde{\Thetaset}$, we construct a linear equation:
\begin{equation}
    \tilde{\Thetaset}= \argmin\nolimits_{\Thetaset} ||f(\X)-\Thetaset f(\NXX)||_2^2.
\end{equation}
We can leverage the least square method to find the solution of this linear equation to be
\begin{equation}
\begin{aligned}
    \tilde{\Thetaset}&=(f^T(\NXX)f(\NXX))^{-1}f^T(\NXX)f(\X), \\
    \tilde{\E} &=f(\X)-\tilde{\Thetaset} f(\NXX),
\end{aligned}
\end{equation}
where $\tilde{\E}=\{\tilde{e}_i\}_{i=1}^N\in \R^N$ denotes the residual errors with the optimal parameter $\tilde{\Thetaset}$, representing those indescribable components in $f(\X)$. Generally, texture regions tend to yield larger residual errors than flat regions due to their low self-description ability. Therefore, we define the texture complexity as
\begin{equation}
    C(\X)=\log_2(1+\frac{1}{N}\sum_{i=1}^N|\tilde{e}_i|).    
\end{equation}
The residual error maps and texture complexity values of four reference point clouds on ICIP2020 are visualized in Fig. \ref{fig:AR_residual}, where the "Longdress" and "Soldier" references have complex patterns
while the "Sarah" and "Ricardo" references have abundant homogenous regions. We can see that: i) the corresponding residual maps accurately reflect the characteristics of different regions in various point clouds; ii) the rank of the texture complexity values is consistent with human perception, which demonstrates the effectiveness of the proposed measurement.

\subsubsection{Texture Masking Compensation}

After quantitatively estimating the texture complexity with the AR model, it is critical to explore how $C(\X)$ influences perceptual distortion. Since Section \ref{sec:texture_masking} has revealed that the MOS-$\rm PSNR_Y$ curves of different references share a similar shape but different horizontal displacements $\Delta$, our objective is to investigate the relationship between $\Delta$ and $C(\X)$, that is, to find $f(\cdot)$ in Eq. \eqref{eq:texture_masking_2}. 

\begin{figure}
    \centering
    \subfigure[]{\includegraphics[width=0.48\linewidth]{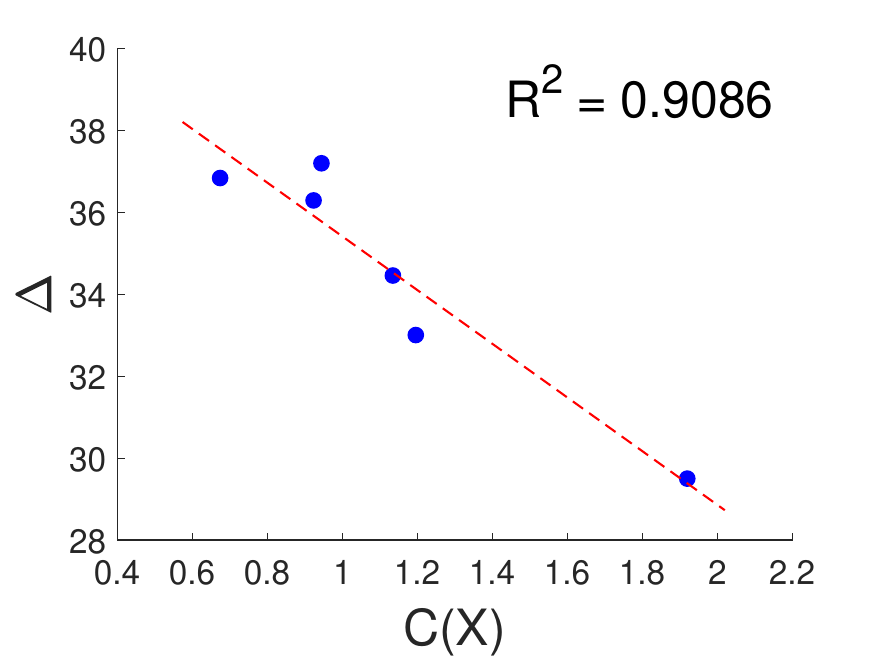}}
    \subfigure[]{\includegraphics[width=0.48\linewidth]{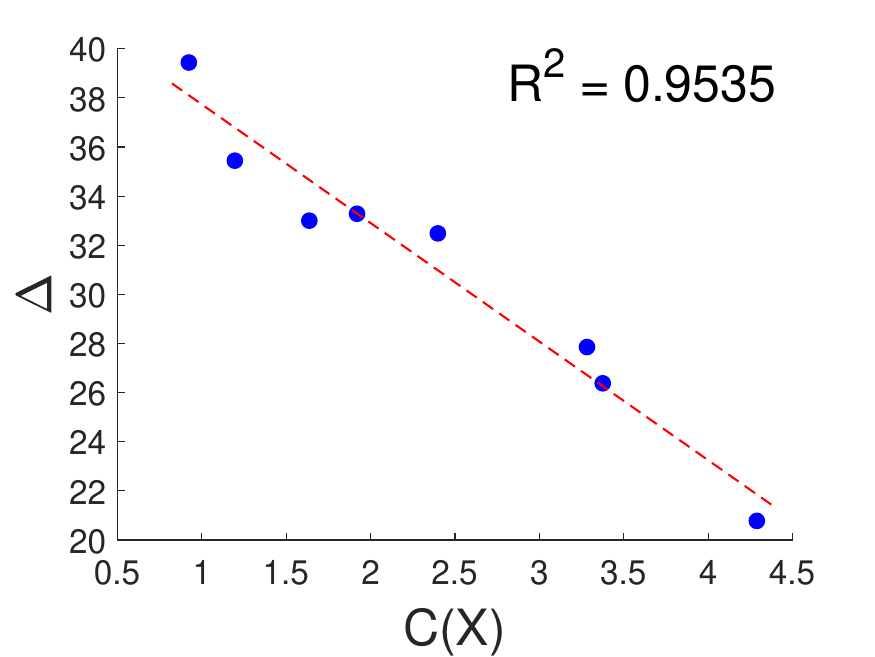}}
    \subfigure[]{\includegraphics[width=0.48\linewidth]{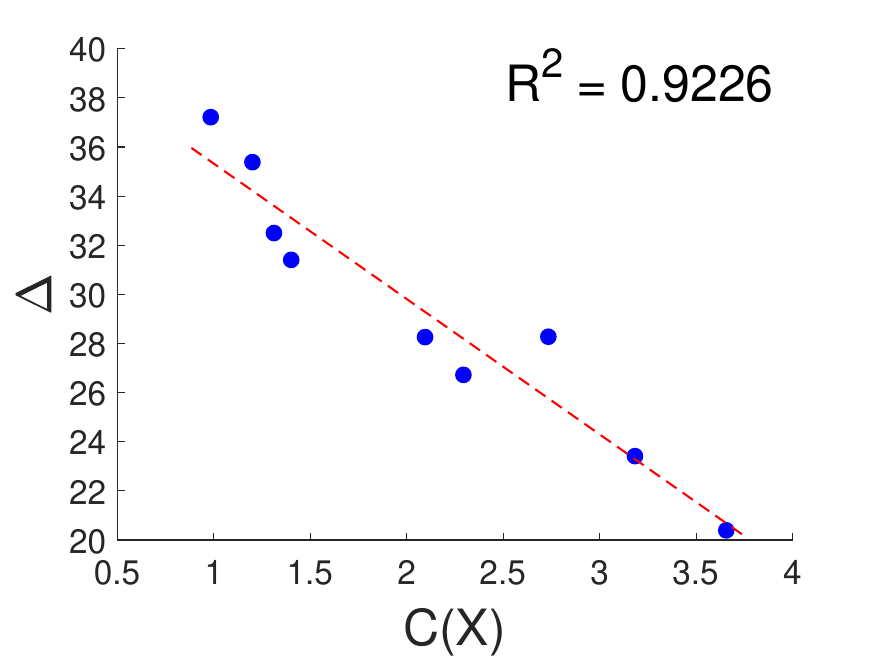}}
    \subfigure[]{\includegraphics[width=0.48\linewidth]{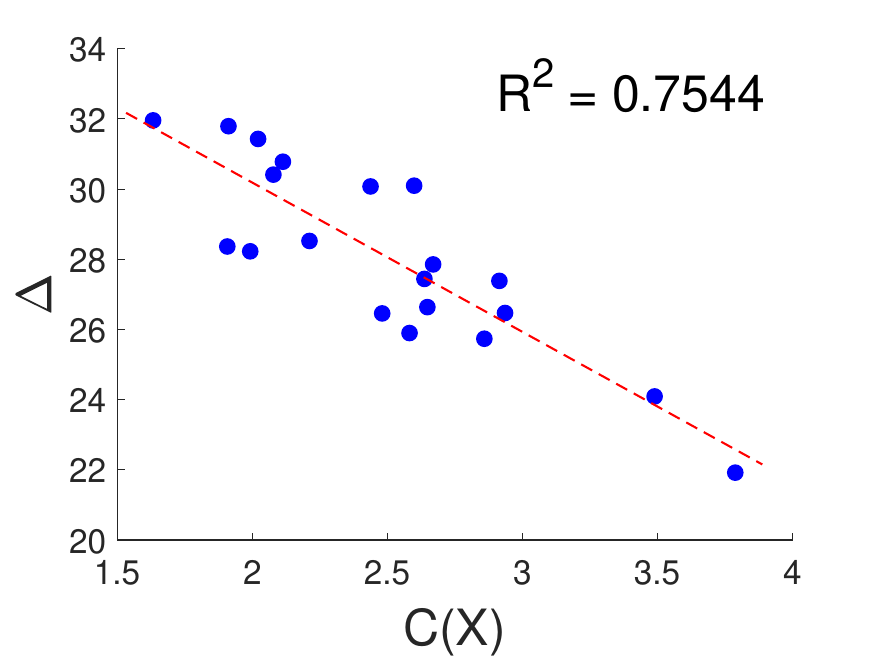}}
    \caption{The relationship between the texture complexity $C(\X)$ and the horizontal displacement $\Delta$ on four databases. (a) ICIP2020. (b) M-PCCD. (c) SJTU-PCQA. (d) WPC. The $R^2$ values of linear fitting are presented.}
    \label{fig:complexity_displacement_plot}
\end{figure}

For this purpose, we illustrate the relationship between
$C(\X)$ and $\Delta$ on multiple databases (ICIP2020, M-PCCD \cite{alexiou2019comprehensive}, SJTU-PCQA \cite{yang2020predicting}, and WPC \cite{liu2022perceptual}) in Fig. \ref{fig:complexity_displacement_plot}. Note that the actual horizontal displacement of the curves is
measured by the intersection of the curves and any horizontal lines such as MOS = 3 or 6 (for databases with MOS = 5 or 10 as the best quality). From Fig. \ref{fig:complexity_displacement_plot}, we observe that $\Delta$ decreases linearly with $C(\X)$ for all databases. Therefore their relationship is derived  as 
\begin{equation}
    \Delta=f(C(\X))=-\alpha\cdot C(X) +\beta,
    \label{eq:complexity_displacement_relation}
\end{equation}
where $\alpha$ and $\beta$ are both positive parameters. Then by substituting Eq. \eqref{eq:complexity_displacement_relation} into Eq. \eqref{eq:texture_masking_2}, we have
\begin{equation}\label{eq:texture_masking_3}
      \psi({\rm PSNR_Y}+\alpha \cdot C(\X) -\beta)=q.
\end{equation}
Considering that $\beta$ is shared by all samples, we finally define the visible difference compensated by the texture complexity as
\begin{equation}\label{eq:DH_evaluation}
    D_H(\X,\Y)=\frac{1}{\Upsilon}({\rm PSNR_Y}+\alpha\cdot C(\X)),
\end{equation}
\znote{where $\Upsilon$ is is a constant parameter which is used to normalize the $D_H$ value into the range of $[0,1]$. For this purpose, we set $\Upsilon=10\log_{10}(255^2)+\alpha\cdot8$. Such a setting is based on the fact that the mean squared error $d_{\X2\Y}$ and $d_{\Y2\X}$ between two point clouds in the $\rm PSNR_Y$ calculation is generally more than 1 and the residual error $|\tilde{e}_i|$ in the complexity calculation is generally less than 255. Similar to \cite{wu2012perceptual}, to avoid infinite $\rm PSNR$, for the rare cases that $d_{\X2\Y}$ and $d_{\Y2\X}$ are less than 1 and even tend to 0, we regard them as having almost perfect quality and simply set $D_H(\X,\Y)=1$ if $\frac{1}{\Upsilon}(\mathrm{PSNR}_Y+\alpha\cdot C(\X))>1$ in the implementation.}

\begin{figure}
    \centering
    \subfigure[]{\includegraphics[width=0.8\linewidth]{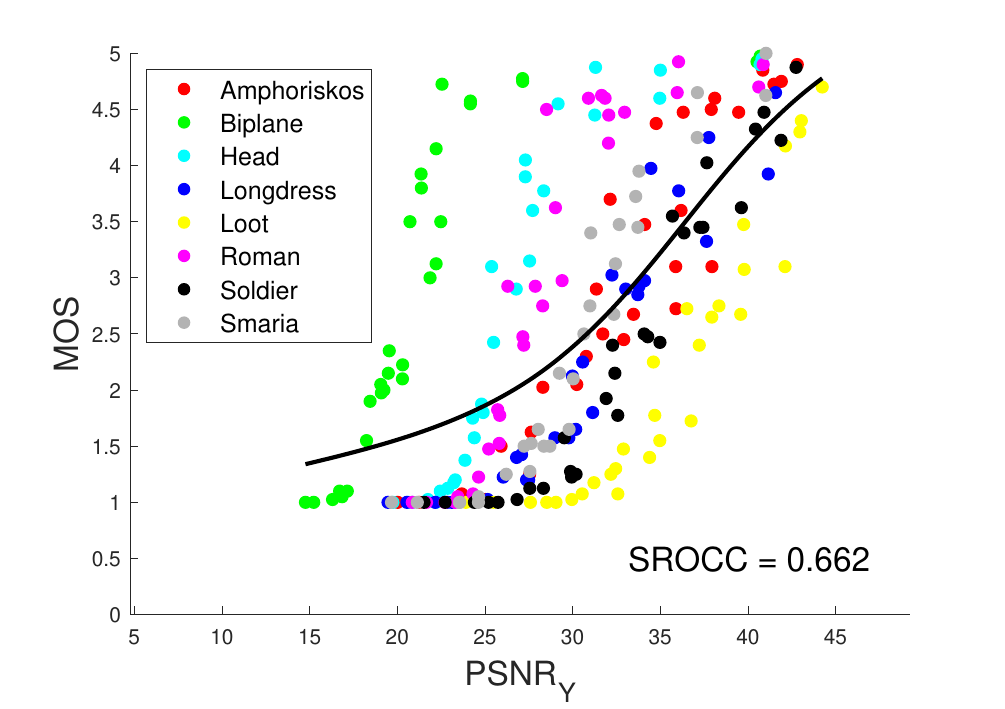}}
    \subfigure[]{\includegraphics[width=0.8\linewidth]{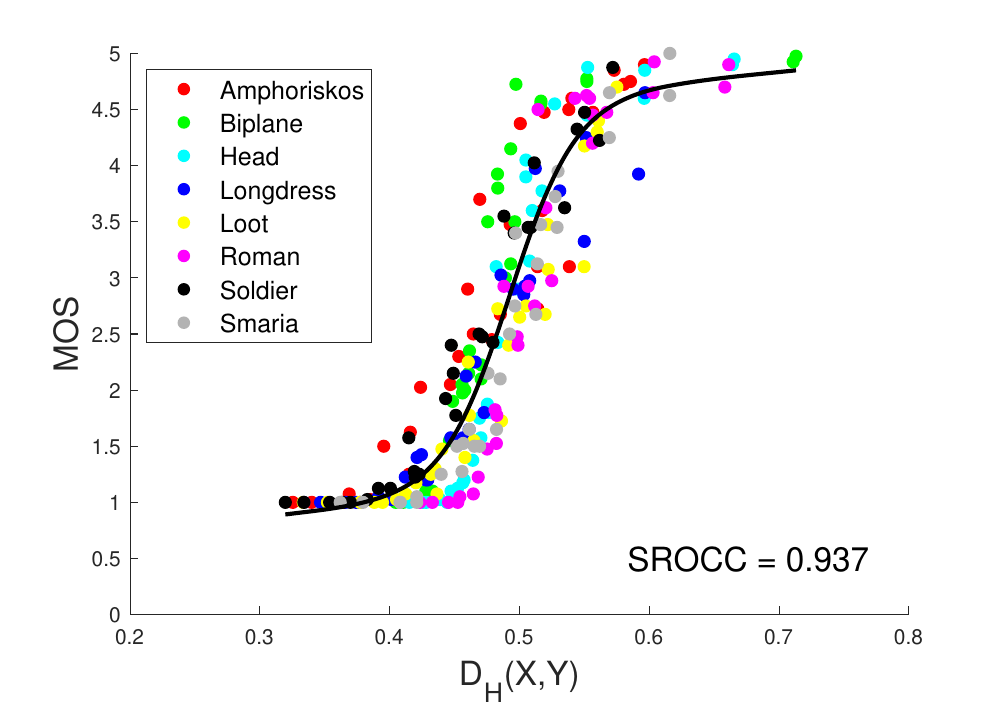}}

    \caption{Scatter plots of MOS versus $\rm PSNR_Y$ and $D_H(\X,\Y)$ on M-PCCD. (a) Plot of MOS versus $\rm PSNR_Y$. (b) Plot of MOS versus $D_H(\X,\Y)$.  }
    \label{fig:MOS_DH_plot}
\end{figure}

To better show the feasibility of the proposed compensation strategy, we present the scatter plots of MOS versus $\rm PSNR_Y$ and $D_H(\X,\Y)$ on the M-PCCD database in Fig. \ref{fig:MOS_DH_plot}, where the 
Spearman rank order correlation coefficients (SROCCs) are also illustrated. Evidently, by adopting the texture complexity as the compensation factor, the scatter points generated by $D_H(\X,\Y)$ are grouped more compactly around the fitted curve and achieve a higher SROCC, which justifies the effectiveness of the proposed compensation method.

\subsection{Appearance degradation Measurement} \label{sec:measure appearance degradation}

\subsubsection{Graph Construction}
To facilitate structure-wise comparison, we initially refer to \cite{zhang2023tcdm} to perform space segmentation for both $\X$ and $\Y$ utilizing a 3D Voronoi diagram. Specifically, we first derive the generating seed set $\S=\{\s_l\}_{l=1}^L\in\R^{L\times6}$ from $\X$ via the farthest point sampling (FPS) \cite{eldar1997farthest}, where the sampling ratio is determined as $L=N/1000$ following \cite{yang2020inferring}. Then, for each seed $\s_l$, we define its corresponding Voronoi cell $VC_l$ as
\begin{equation}\label{eq:VC}
    VC_l = \{\p\in\R^3|||\p-\s_l^O||_2^2\leqslant ||\p-\s_k^O||_2^2, if\quad k\neq l\}.
\end{equation}
The Voronoi cell $VC_l$ denotes one convex polyhedron in which every point is much closer to  $\s_l$ than to other seeds.

Next we collect the local patch pair bounded in $\X$ and $\Y$ by the $VC_l$ as
\begin{equation}
\begin{aligned}
    \X_l&=\{\x_i|\x_i^O\in VC_l\}\in \R^{N_l\times6},\\
    \Y_l&=\{\y_i|\y_i^O\in VC_l\}\in \R^{M_l\times6},
\end{aligned}
\end{equation}
where $N_l$ and $M_l$ denote the point number of two local patches. Consequently, we obtain $L$ local patch pairs. Note that all patches in $\X$ or $\Y$ are non-overlapping and no point is omitted during the above process.

To model point clouds via the spectral graph theory, we construct the graph over each local patch based on the affinity of geometric distance among points. Taking points in $\X_l$ as an instance, we assign the edge weight $w_{\x_i,\x_j}$ between two vertices $\x_i$ and $\x_j$ given by
\begin{equation}
	w_{\x_i,\x_j}=\left\{
	\begin{aligned}
		\exp{(- \frac{||\x_i^O-\x_j^O||_2^2 }{\sigma^2}}), \x_j\in\NxXl \\
		0,\quad otherwise,
	\end{aligned}
	\right. 
\label{eq:edge_weight}
\end{equation}
where $\NxXl \subset\X_l$ denotes $\x_i$'s $K_2$ closest points in $\X_l$ and $K_2=10$ is determined empirically. \znote{$\sigma^2$ represents the geometry variance and is defined as $\sigma^2=(\sum\nolimits_{\x_i\sim\x_j}||\x_i^O-\x_j^O||_2^2))/N_e$,
where $\x_i\sim \x_j$ means vertices $\x_i$ and $\x_j$ are connected in the graph; $N_e$ denotes the number of edges.} $\Y_l$ can be modeled in a similar way. Finally, we represent each local patch pair using graphs involving points and their neighbor connections, noted as $\G_{\X_l}$ and $\G_{\Y_l}$, whose Laplacian matrices are marked as $\L_{\X_l}$ and $\L_{\Y_l}$.

\subsubsection{Geometry Appearance Degradation Evaluation}
Geometry deformations (e.g., geometry noise, octree pruning, and down-sampling) cause the change in surface smoothness, which can be well measured by the signal variation on graphs. Specifically, by defining the 3D coordinates of $\X_l$ as graph signals on $\G_{\X_l}$, their variation on the graph can be described via \textit{graph smoothness}. Denoting $f(\X_l)$ as 
x-axis coordinates, the graph smoothness over the x-axis can be defined as 

\begin{equation}
\begin{aligned}
    S_{\X_l,X}&=f^T(\X_l)\L_{\X_l} f(\X_l)\\
    &=\sum\nolimits_{\x_i\sim \x_j}w_{\x_i,\x_j}(f(\x_i)-f(\x_j))^2\\
    &=\sum\nolimits_{i=1}^{N_l}\lambda_i\hat{f}^2(\x_i),
\end{aligned}
\end{equation}
where $\lambda_i$ is the $i$-th eigenvalue of $\L_{\X_l}$ and $\hat{f}(\x_i)$ is the $i$-th GFT coefficient. The graph smoothness over the y- and z-axes (denoted by $S_{\X_l,Y}$ and $S_{\X_l,Z}$) can be computed in a similar way. From a spectral standpoint, $\hat{f}^2(\x_i)$ represents the energy associated with the $i$-th
graph frequency for geometry information. Considering $\lambda_i$ is organized in ascending order, a small graph smoothness means that most of the signal energy is occupied by the low-frequency components, and changes in graph smoothness can be an effective indicator of geometric alterations.

Given the variation in the number of points among patches, we subsequently normalize the graph smoothness as $\bar{S}_{\X_l,X}=\frac{1}{N_l}S_{\X_l,X}$. The local index of the geometry appearance degradation over x-axis is then calculated as 
\begin{equation}
   F^S_{l,X}=\frac{2\bar{S}_{\X_l,X}\cdot \bar{S}_{\Y_l,X}+T}{(\bar{S}_{\X_l,X})^2+(\bar{S}_{\Y_l,X})^2+T},
\end{equation}
where $T$ is set as $10^{-6}$ in our metric. $F^S_{l,Y}$ and $F^S_{l,Z}$ can be derived in a similar way.

Finally, to assess the global geometry appearance degradation, we average all local indices in three geometry channels, i.e.,

\begin{equation}
    D_L^O(\X,\Y)=\frac{1}{3L}\sum\nolimits_{l=1}^L(F_{l,X}^S+F_{l,Y}^S+F_{l,Z}^S).
\end{equation}

\subsubsection{Texture appearance Degradation Evaluation} 
Receptive fields in the primary visual cortex can be characterized as being localized and band-pass \cite{fournier2011adaptation}, comparable to the basis of wavelet. Therefore, from the perspective of HVS perception, the wavelet features are appropriate in the quality evaluation. To extend the wavelet analysis to irregular point cloud data, we propose to measure the texture appearance degradation using the SGWT.

\begin{figure}
    \centering
    \includegraphics[width=\linewidth]{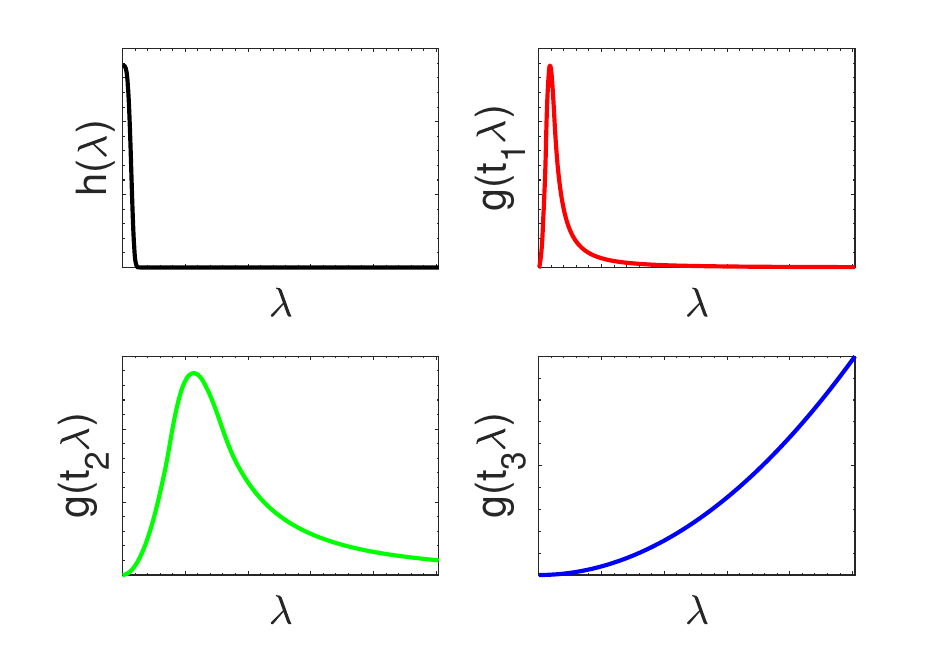}
    \caption{\znote{Illustration of wavelet filters (a low-pass filter $h(\lambda)$ and three band-pass filters $g(t_1\lambda)-g(t_3\lambda)$) presented in the graph spectral domain. The horizontal axis represents the graph frequency, and the vertical axis denotes the amplitude of the filter response.}}
    \label{fig:wavelet_filter}
 
\end{figure}

A key part of the SGWT utilization is the choice of the wavelet function $g(\cdot)$ and the scaling function $h(\cdot)$ (cf. Eq. \eqref{eq:wavelet_function} and \eqref{eq:scaling_function}). Here,  we refer to \cite{hammond2011wavelets} to choose the two functions, that is
\begin{equation}\label{eq:wavelet_filter}
\begin{aligned}
        g(\lambda)&=\left\{
	\begin{aligned}
		&\lambda^2,  &\lambda<1 \\
		&\lambda^3-6\lambda^2+11\lambda-5,  &1\leq\lambda\leq2\\
        &\lambda^{-2},  &\lambda>2,
	\end{aligned}
	\right. \\
    h(\lambda)&=\gamma \exp{(-(\frac{\lambda}{0.6\lambda_{min}})^4)}.     
\end{aligned}
\end{equation}
According to \cite{hammond2011wavelets}, $\gamma$ is set to make $h(0)$ equal the maximum value of $g(\cdot)$, and $\lambda_{min}$ is determined by the upper bound $\lambda_{max}$ of the graph frequency, i.e, $\lambda_{min}=\lambda_{max}/20$. The scale parameters $t_c$ are chosen to be logarithmically equispaced between the minimum and maximum scales $t_1=2/\lambda_{max}$ and
$t_C=2/\lambda_{min}$. In our work, we choose $C=3$, thus obtaining three band-pass filters $\{g(t_c\lambda)\}_{c=1}^3$ and one low-pass filter $h(\lambda)$. Fig. \ref{fig:wavelet_filter} illustrates an example of the four filters in the graph spectral domain. Apparently, $h(\lambda)$ denotes the low-frequency components, while $g(t_{1}\lambda)$-$g(t_{3}\lambda)$ characterize signals from distinct band-pass sub-bands, which provides us with a multi-resolution perspective on PCQA.

\begin{figure}
    \centering
    \subfigure[]{\includegraphics[width=\linewidth]{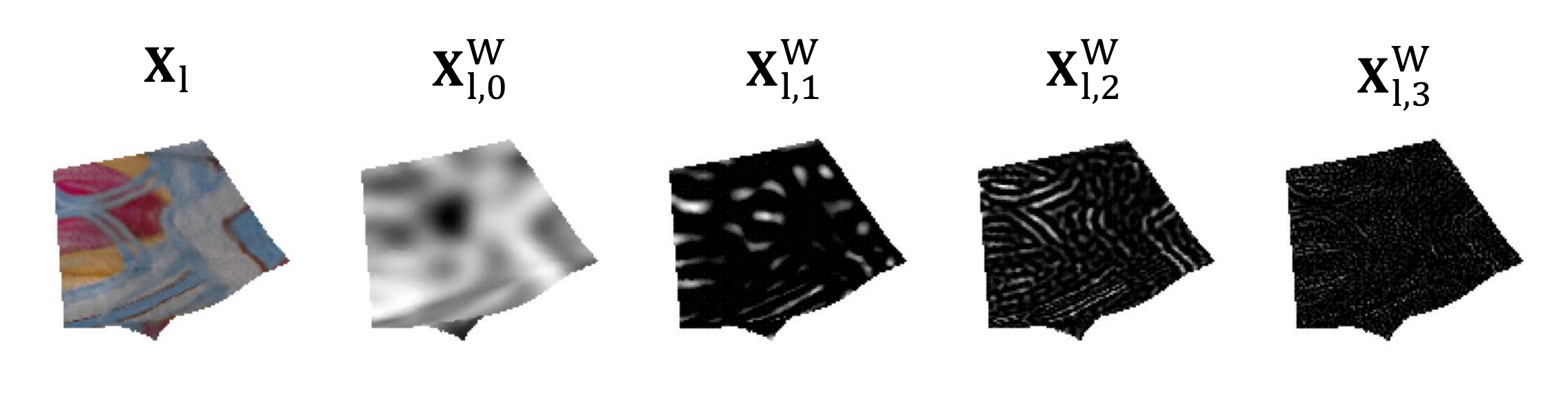}}
    \subfigure[]{\includegraphics[width=\linewidth]{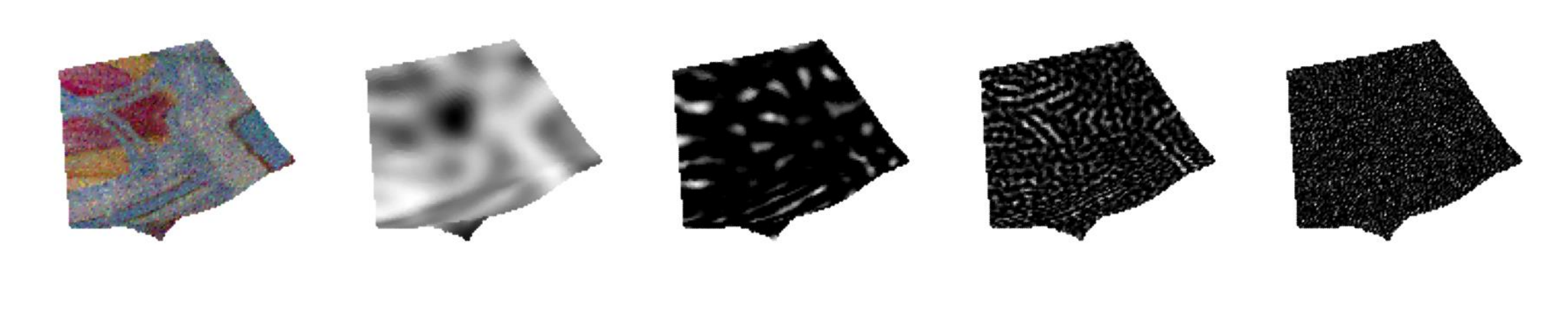}}
    \subfigure[]{\includegraphics[width=\linewidth]{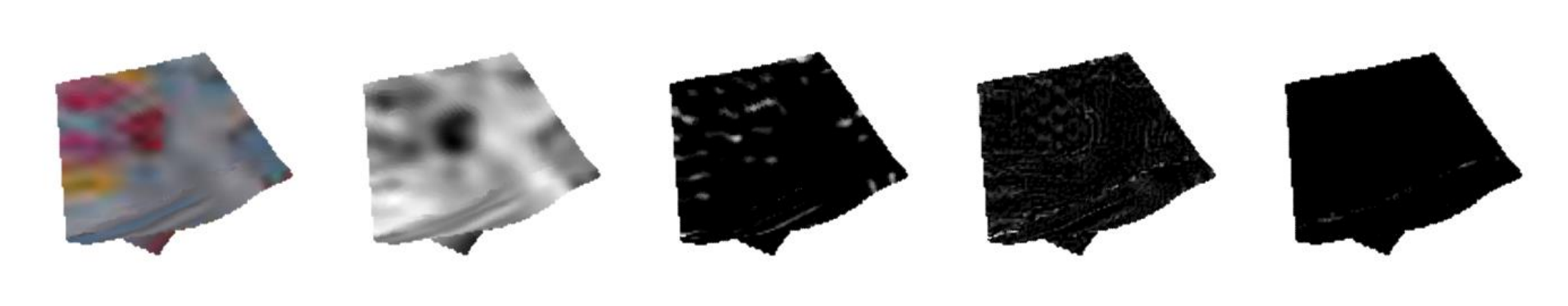}}
    \caption{Illustration of graph wavelet sub-bands on patches with different distortions. (a) Reference; (b) CN. (c) V-PCC. Best viewed when zoomed in. }
    \label{fig:wavelet_sample}
\end{figure}

We further use the filters in Eq. \eqref{eq:wavelet_filter} to decompose the luminance channel of each patch into four sub-bands based on Eq. \eqref{eq:wavelet_function} and \eqref{eq:scaling_function} and note the sub-band wavelet coefficients of $\X_l$ as $\X_l^W=\{\X_{l,c}^W\}_{c=0}^3$, where $\X_{l,c}^W=\{\x_{i,c}^W\}_{i=1}^{N_l}\in\R^{N_l}$ denote the wavelet coefficients on the $c$-th sub-band. Fig. \ref{fig:wavelet_sample} illustrates several examples of wavelet decomposition. In Fig. \ref{fig:wavelet_sample}, the first column of three sub-figures (a)-(c) depicts one local patch of the "Longdress" sample and its two distorted versions caused by color noise (CN) and video-based point cloud compression (V-PCC), and the last four columns show their associated wavelet sub-bands. From Fig. \ref{fig:wavelet_sample}, we have the following observations. First, the sub-bands reflect hierarchical information of point cloud texture. Specifically, $\X^W_{l,0}$ characterizes the coarse structure of the original sample, while $\X^W_{l,1}$-$\X^W_{l,3}$ respectively identify the detail information such as edges. Second, CN mostly disturbs the high-frequency sub-band (i.e., the fourth column in (b)) but almost does not deform the low-frequency sub-band. This is because CN locates multiple isolated points and does not influence the coarse structure. Third,  V-PCC almost disturbs all sub-bands. This is due to the fact that the quantization of attributes in V-PCC affects the texture distribution, leading to severe texture degradation.

\begin{figure*}
    \centering
    \subfigure[]{\includegraphics[width=0.8\linewidth]{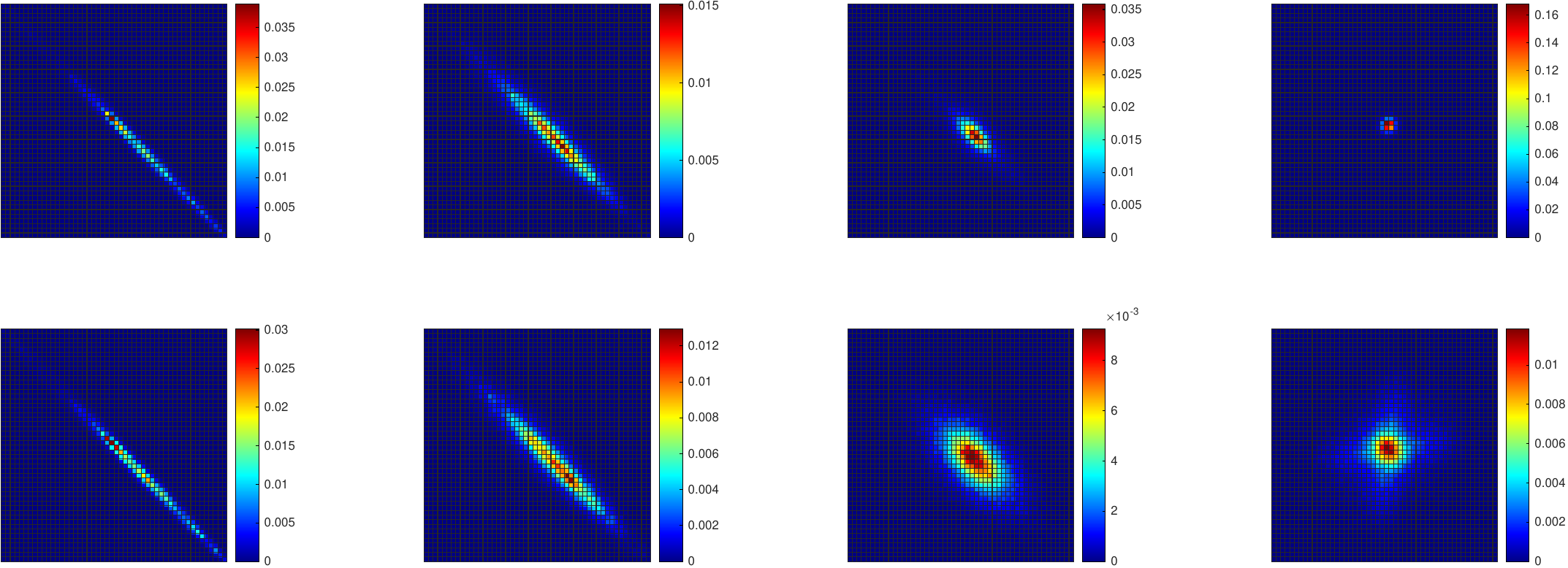}}
    \subfigure[]{\includegraphics[width=0.8\linewidth]{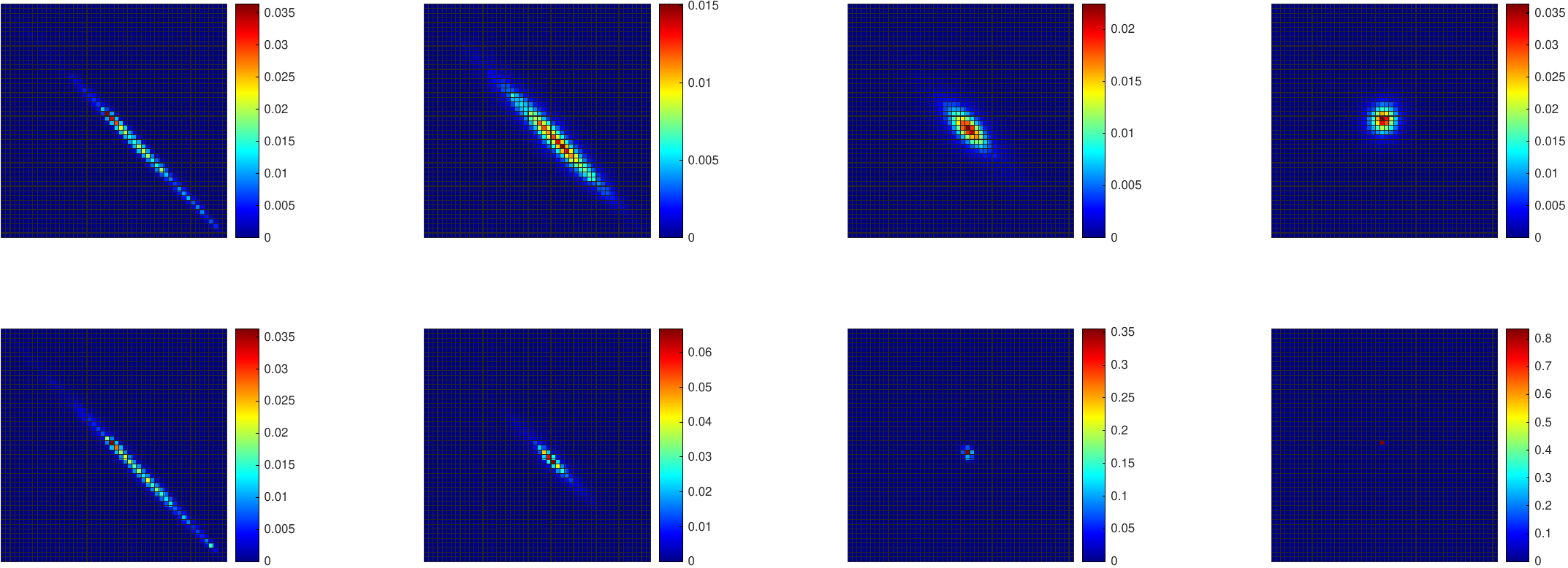}}
    \caption{Illustration of normalized WCM on patches with different distortions. (a) CN. (b) V-PCC. The top row of each sub-figure corresponds to the normalized WCM of the reference patch in Fig. \ref{fig:wavelet_sample} (a). The second rows of the two sub-figure respectively correspond to the normalized WCM of the distorted patches in Fig. \ref{fig:wavelet_filter} (b) and (c). \znote{Note that there are some differences between the WCMs of the reference in (a) and (b) because the quantized range are different for two types of distortion.} }
    \label{fig:comatrix_sample}
\end{figure*}

To gague the texture appearance degradation, we use the weighted co-occurrence matrices (WCM) of wavelet coefficients to reflect "where and how" points are composed together \cite{lu2012saliency}. Specifically, We first quantize each two corresponding wavelet sub-bands (i.e., $
\X^W_{l,c}$ and $\Y^W_{l,c}$) into a limited bin number $N_b=50$  within the same range $[b_{min},b_{max}]$, where $b_{min}=\mathrm{min}([\X^W_{l,c};\Y^W_{l,c}])$ and $b_{max}=\mathrm{max}([\X^W_{l,c};\Y^W_{l,c}])$. Denoting the 
quantized value set as $\b=\{b_m\}_{m=1}^{N_b}$, the weighted co-occurrence matrix of $\X^W_{l,c}$ is defined as
\begin{equation}\label{eq:co-occurrence_matrix}
\begin{aligned}
    \Gamma_{\X_l,c}(m,n)&= \sum\nolimits_{\x_i\sim \x_j}\rho_{m,n}(\x_i,\x_j)w_{\x_i,\x_j},  \\
    \rho_{m,n}(\x_i,\x_j)&=\left\{
	\begin{aligned}
		&1,  &(\x^W_{i,c},\x^W_{j,c})=(b_m,b_n) \mathrm{or} (b_n,b_m) \\
		&0,  &\mathrm{otherwise},\\
        \end{aligned}
        \right.     
\end{aligned}
\end{equation}
Eq. \eqref{eq:co-occurrence_matrix} builds the WCM as follows. For each connected point pair $(\x_i,\x_j)$ on the graph $\G_{\X_l}$, the element of the WCM $\Gamma_{\X_l,c}(m,n)$ is incremented by the edge weight $w_{\x_i,\x_j}$ if the quantized wavelet coefficients of the two points form the pair $(b_m,b_n)$. In other words,
the closer the point pairs with quantized wavelet values $(b_m, b_n)$ are, the larger the corresponding element value $\Gamma_{\X_l,c}(m,n)$. 
Generally, the diagonal elements of the WCM indicate the distribution of low-contrast point pairs, while the elements far from the diagonal indicate the distribution of high-contrast point pairs.

\begin{figure}
    \centering
    \subfigure[]{\includegraphics[width=0.8\linewidth]{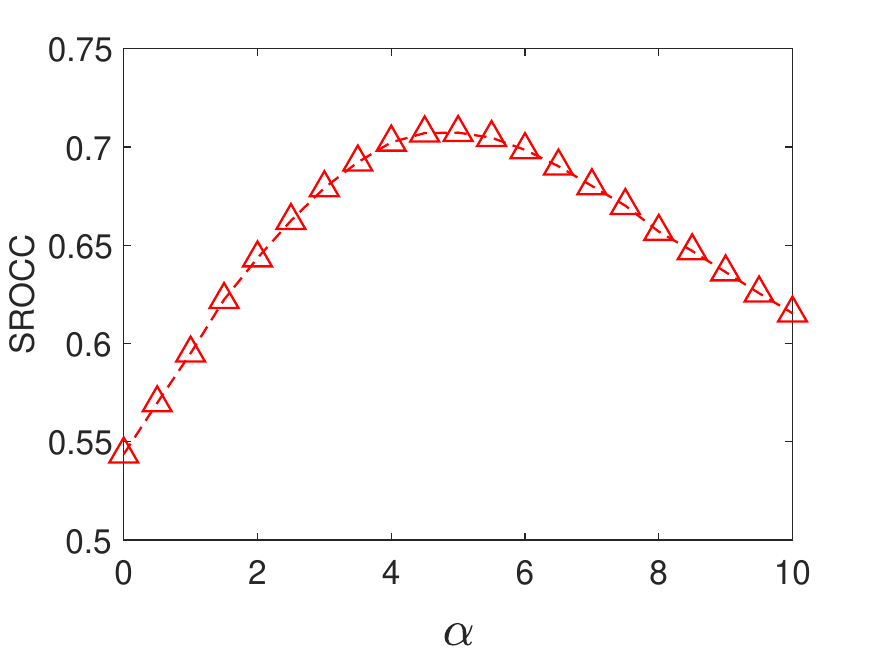}}
    \subfigure[]{\includegraphics[width=0.8\linewidth]{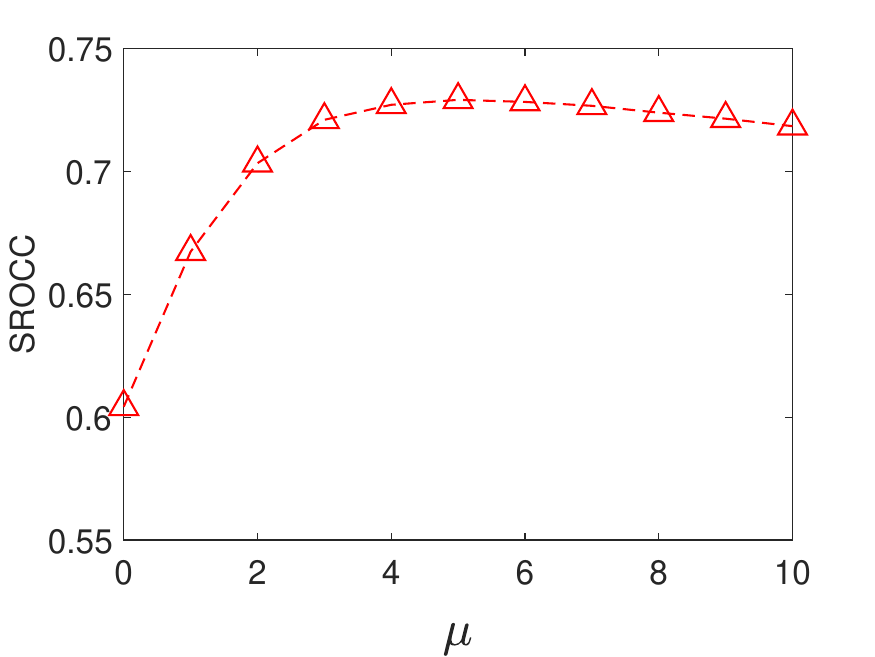}}
    
    \caption{Impact of different parameter settings. (a) Impact of $\alpha$. (b) Impact of $\mu$.}
    \label{fig:parameter_selection}
    \vspace{-0.4cm}
\end{figure}

The WCM is further normalized as $\bar{\Gamma}_{\X_l,c}=\Gamma_{\X_l,c}/\sum_m\sum_n\Gamma_{\X_l,c}(m,n)$. Fig. \ref{fig:comatrix_sample} presents examples of the normalized WCM corresponding to the patches shown in Fig. \ref{fig:wavelet_sample}. In the figure, the first rows of two sub-figures (a) and (b) denote the WCM of the reference patch on the four wavelet sub-bands (i.e.,  $\bar{\Gamma}_{\X_l,0}$-$\bar{\Gamma}_{\X_l,3}$) and the second rows denote the WCM of distorted patches impaired by CN and V-PCC (i.e., $\bar{\Gamma}_{\Y_l,0}$-$\bar{\Gamma}_{\Y_l,3}$).
\znote{Note that there are some differences between the WCMs of the reference in Fig. \ref{fig:comatrix_sample} (a) and (b) because the quantized range $[b_{min},b_{max}]$ is different for the two types of distortion.} From the figure, we have the following observations. First, the energy distribution of the WCM associated with the low-frequency sub-band predominantly concentrates along the diagonal. This is due to the reason that the low-frequency sub-band discards most high-contrast information such as edges and noises. Second, in the context of the high-frequency sub-bands, the energy distribution of the WCM associated with CN exhibits a more dispersed pattern, whereas the WCM corresponding to V-PCC concentrates significantly around the origin. The results can be ascribed to the fact that CN generates many outliers that form high-contrast point pairs with neighbors, while V-PCC blurs much detailed information. Based on this instance, we can see that the proposed WCM can effectively capture and distinguish various texture deformations.

We then measure the local appearance discrepancy by comparing the co-occurrence matrices as follows,
\begin{equation}
    F^W_{l,c}=\mathrm{PLCC}(\bar{\Gamma}_{\X_l,c},\bar{\Gamma}_{\Y_l,c}),
\end{equation}
where $\mathrm{PLCC}(\cdot)$ measures the Pearson linear correlation coefficient between two matrices. Finally, the texture appearance degradation is calculated as 
\begin{equation}
    D^I_{L}(\X,\Y)=\frac{1}{L(C+1)}\sum\nolimits_{l=1}^L\sum\nolimits_{c=0}^CF^W_{l,c}.
\end{equation}

\subsubsection{Appearance Degradation Fusion}

After deriving the geometry and texture appearance degradation respectively, we simply multiply the two components (cf.  Eq. \eqref{eq:apperance_fusion}) to obtain the quality index $D_L(\X,\Y)$ for low-quality samples.

\subsection{Adaptive Combination}
The previous sections presented two measurements targeted for different distortion levels: $D_H(\X,\Y)$ for high-quality level and $D_L(\X,\Y)$ for low-quality level. The hybrid strategy requires an appropriate $\omega$ in Eq. \eqref{eq:hybrid_model} to make the metric dynamically assign weights to the two components in different scenarios. Inspired by \cite{larson2010most}, we define the final objective metric as 
\begin{equation}\label{eq:final_index}
    \begin{aligned}
        D(\X,\Y)&=D_H^{1-\omega(D_H)}(\X,\Y)\cdot D_L^{\omega(D_H)}(\X,\Y),\\
        \omega(D_H)&=\frac{1}{1+\mu D_H(\X,\Y)},
    \end{aligned}
\end{equation}
where $\mu$ is the predetermined parameter optimized based on a subjective database \cite{liu2021reduced}. The weight $\omega(D_H)$ is incremented with the decrease of $D_H(\X,\Y)$ (note that a higher $D_H(\X,\Y)$ indicates better quality), which helps the metric emphasize $D_L(\X,\Y)$ for low-quality samples and vice versa.

\section{Experiment}\label{sec:experiment}

This section evaluates the proposed metric and other state-of-the-art metrics for point cloud quality prediction, using five publicly accessible point cloud databases.
\subsection{Databases and Evaluation Protocols}
Five benchmark databases are used for experimental evaluation, including SJTU-PCQA \cite{yang2020predicting}, WPC \cite{liu2022perceptual}, LS-PCQA \cite{liu2020point}, M-PCCD \cite{alexiou2019comprehensive} and ICIP2020 \cite{perry2020quality}. In SJTU-PCQA, 7 types of distortion, namely, octree-based compression (OT), color noise (CN), geometry Gaussian noise (GGN), Down-sampling (DS), and three superimposed distortions, i.e., DS and CN (D+C), DS and GGN (D+G), CN and GGN (C+G) are introduced to obtain 378 distorted samples. WPC contains 20 reference point clouds and 740 distorted point clouds generated from the references under five types of distortions, including DS, Gaussian noise contamination (GNC), G-PCC(Trisoup), G-PCC(Octree), and V-PCC. LS-PCQA is a large database with 85 reference point clouds and 930 distorted point clouds impaired by 31 types of distortion. M-PCCD contains eight reference point clouds and 232 distorted samples impaired by five compression modes, including V-PCC and four variations of G-PCC. ICIP2020 is comprised of six references and 90 distorted samples, and the distortion results from three compression modes, including V-PCC, G-PCC(Trisoup), and G-PCC(Octree).

\begin{table*}
\caption{OVERALL PERFORMANCE COMPARISON OF DIFFERENT FR-PCQA METRICS ON FIVE DATABASES}
\label{tab:overall_performance}
\resizebox{\linewidth}{!}{
\setlength{\tabcolsep}{2.5pt}
\renewcommand\arraystretch{1.2}
\begin{tabular}{l|ccc|ccc|ccc|ccc|ccc|cc}
\toprule
\multirow{2}{*}{metrics} & \multicolumn{3}{c|}{SJTU-PCQA\cite{yang2020predicting}} & \multicolumn{3}{c|}{WPC\cite{liu2022perceptual}} & \multicolumn{3}{c|}{LS-PCQA \cite{liu2020point}} & \multicolumn{3}{c|}{M-PCCD\cite{alexiou2019comprehensive}} & \multicolumn{3}{c|}{ICIP2020\cite{perry2020quality}} &\multicolumn{2}{c}{Avg Rank}  \\ \cmidrule{2-18}
& PLCC & SROCC & RMSE & PLCC & SROCC & RMSE & PLCC & SROCC & RMSE & PLCC & SROCC & RMSE & PLCC & SROCC & RMSE  &PLCC &SROCC
\\ \midrule
PSNR  &0.462	&0.437	&2.152	&0.306	&0.279	&21.822	&0.468	&0.475	&0.735	&0.485	&0.422	&1.19	&0.669	&0.655	&0.844	&15.4	&14.6

\\
SSIM \cite{wang2004image} &0.579	&0.540	&1.979	&0.367	&0.291	&21.321	&0.463	&0.461	&0.737	&0.537	&0.456	&1.148	&0.702	&0.689	&0.809	&14.4	&14.0
\\
MS-SSIM \cite{wang2003multiscale} &0.621	&0.566	&1.902	&0.384	&0.316	&21.168	&0.537	&0.512	&0.702	&0.622	&0.584	&1.066	&0.720	&0.695	&0.789	&11.4	&11.2
\\
GMSD \cite{xue2013gradient} &0.635	&0.570	&1.876	&0.343	&0.298	&21.536	&0.486	&0.470	&0.727	&0.574	&0.540	&1.114	&0.724	&0.698	&0.784	&13.0	&12.2
\\
FSIM \cite{zhang2011fsim} &0.575	&0.521	&1.985	&0.304	&0.232	&21.839	&0.493	&0.479	&0.724	&0.579	&0.532	&1.109	&0.741	&0.709	&0.763	&13.2	&13.0 \\

VIF \cite{sheikh2006image} &0.679	&0.638	&1.783	&0.429	&0.415	&20.704	&0.390	&0.402	&0.766	&0.647	&0.635	&1.038	&0.802	&0.807	&0.678	&12.0	&11.4
\\ \midrule
MSE-p2po \cite{MPEGSoft} &0.877	&0.791	&1.166	&0.578	&0.566	&18.708	&0.487	&0.302	&0.727	&0.778	&0.797	&0.855	&0.888	&0.878	&0.522 &7.2 &8.8
\\
Haussdorf-p2po \cite{MPEGSoft} &0.742	&0.681	&1.628	&0.398	&0.258	&21.028	&0.403	&0.269	&0.761	&0.593	&0.366	&1.095	&0.601	&0.542	&0.908 &13.0 &15.6

\\
MSE-p2pl \cite{MPEGSoft} &0.753	&0.676	&1.596	&0.488	&0.446	&20.013	&0.444	&0.287	&0.745	&0.815	&0.836	&0.788	&0.913	&0.915	&0.463 &8.2 &9.6
\\
Haussdorf-p2pl\cite{MPEGSoft} &0.737	&0.670	&1.639	&0.383	&0.315	&21.171	&0.401	&0.269	&0.762	&0.571	&0.507	&1.117	&0.649	&0.602	&0.865 &14.2 &14.6

\\
$\rm PSNR_{YUV}$ \cite{MPEGSoft} &0.652	&0.646	&1.841	&0.551	&0.536	&19.132	&0.497	&0.476	&0.722	&0.654	&0.660	&1.029	&0.868	&0.867	&0.564 &9.0 &8.8

\\ 
PCQM \cite{meynet2020pcqm} &0.860	&0.847	&1.237	&0.751	&0.743	&15.132	&0.346	&0.321	&0.780	&0.900	&0.915	&0.594	&\textbf{0.969}	&\textbf{0.970}	&\textbf{0.280}
&6.6 &5.6
\\
GraphSIM \cite{yang2020inferring} &0.856	&0.841	&1.071	&0.694	&0.680	&16.498	&0.355	&0.332	&0.778	&0.932	&\textbf{0.945}	&0.493	&0.890	&0.872	&0.518 &8.0 &6.6
\\
MS-GraphSIM \cite{zhang2021ms} &0.897	&0.874	&1.071	&0.717	&0.707	&15.97	&0.432	&0.404	&0.750	&0.916	&0.930	&0.545	&0.906	&0.895	&0.481 &6.0 &5.6
\\
pointSSIM \cite{alexiou2020towards} &0.725	&0.704	&1.672	&0.510	&0.454	&19.713	&0.291	&0.157	&0.796	&0.926	&0.918	&0.514	&0.904	&0.865	&0.486 &9.8 &10.0
\\
MPED \cite{yang2021mped} &0.896	&0.884	&1.076	&0.700	&0.678	&16.374	&\textbf{0.648}	&\textbf{0.601}&\textbf{0.634}	&0.824	&0.849	&0.770	&0.964	&0.952	&0.302
&\bf 4.0 &4.0
\\ 

TCDM \cite{zhang2023tcdm} &\textbf{0.930}	&\textbf{0.910}	&\textbf{0.891}	&\textbf{0.807}	&\textbf{0.804}	&\textbf{13.525}	&0.433	&0.408	&0.75	&\textbf{0.936}	&0.944	&\textbf{0.479}	&0.942	&0.935	&0.382 &\bf 4.0 &\bf 3.8
	\\ \midrule
PHM  &\textbf{0.907}	&\textbf{0.888}	&\textbf{1.021}	&\textbf{0.839}	&\textbf{0.833}	&\textbf{12.466}	&\textbf{0.615}	&\textbf{0.589}	&\textbf{0.656}	&\textbf{0.945}	&\textbf{0.948}	&\textbf{0.445}	&\textbf{0.969}	&\textbf{0.958}	&\textbf{0.279} &\textbf{1.6} & \textbf{1.6}
\\ \bottomrule

\end{tabular}}
\end{table*}

To ensure the consistency between MOS and objective predictions from various metrics, we first map the objective predictions of different metrics to the same dynamic range following the recommendations suggested by the video quality experts group (VQEG)  \cite{video2003final}, i.e.,
\begin{equation}\label{eq:logic_fit}
    \psi(x) = \beta_1(\frac{1}{2}-\frac{1}{1+\exp{\beta_2(x-\beta_3)}})+\beta_4x+\beta_5,
\end{equation}
where  $\beta_1$, $\beta_2$, $\beta_3$, $\beta_4$, and $\beta_5$ are the parameters fitted by minimizing the sum of squared errors.

Then, the Pearson linear correlation coefficient (PLCC), the Spearman rank order correlation coefficient (SROCC), and the root mean square error (RMSE) are utilized to evaluate the performance of different metrics, which indicate the linearity, monotonicity, and accuracy, respectively. A better objective PCQA measure is expected to have higher SROCC and PLCC while lower RMSE values.

\subsection{Parameter Determination}
Before conducting comprehensive experiments, several parameters should be determined, including the modulation parameter $\alpha$ in Eq. \eqref{eq:DH_evaluation} and the parameter $\mu$ in Eq. \eqref{eq:final_index}. We determine these parameters based on another subjective database \cite{liu2021reduced} and SROCC is used as the criterion for the determination.

\begin{itemize}
    \item Impact of $\alpha$.
    Fig. \ref{fig:parameter_selection} (a) shows the performance of $D_H(\X,\Y)$ with respect to different $\alpha$. We can see that as the $\alpha$ value increases, the performance initially improves and subsequently decreases quickly. This is because too large $\alpha$ overestimates the influence of texture masking. Finally, $\alpha=4.5$ is determined to be used in our experiments.
    \item Impact of $\mu$.
    Fig. \ref{fig:parameter_selection} (b) shows the performance of the metric with different settings of $\mu$. It is observed that too small or too large parameters impair the model performance. The main reason is that these parameters result in a skewed emphasis towards one component while disregarding the other. Finally, we choose $\mu=5$ in the paper.
    
\end{itemize}

\begin{figure}
    \centering
    \subfigure[]{\includegraphics[width=0.48\linewidth]{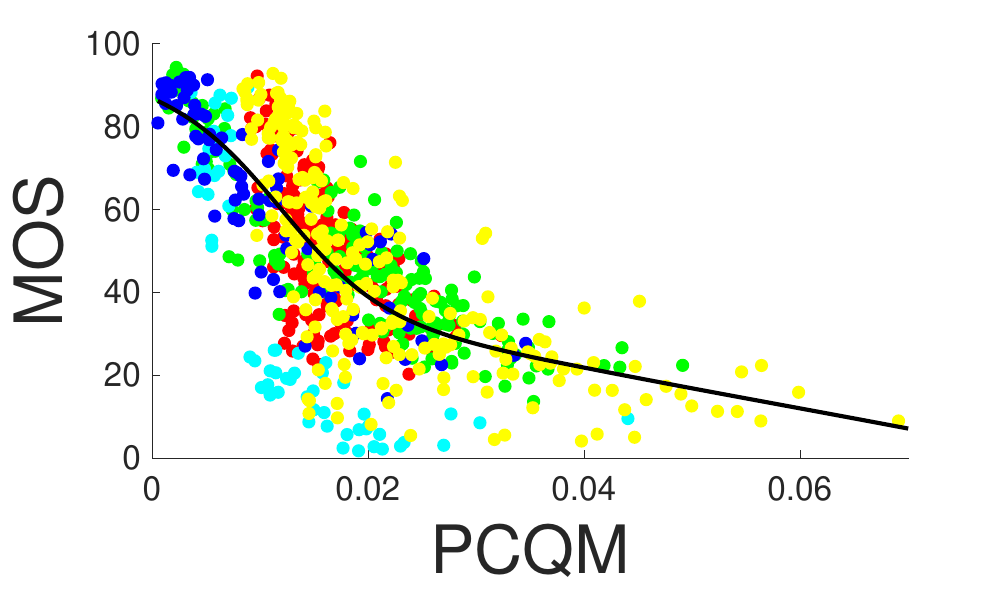}}
    \subfigure[]{\includegraphics[width=0.48\linewidth]{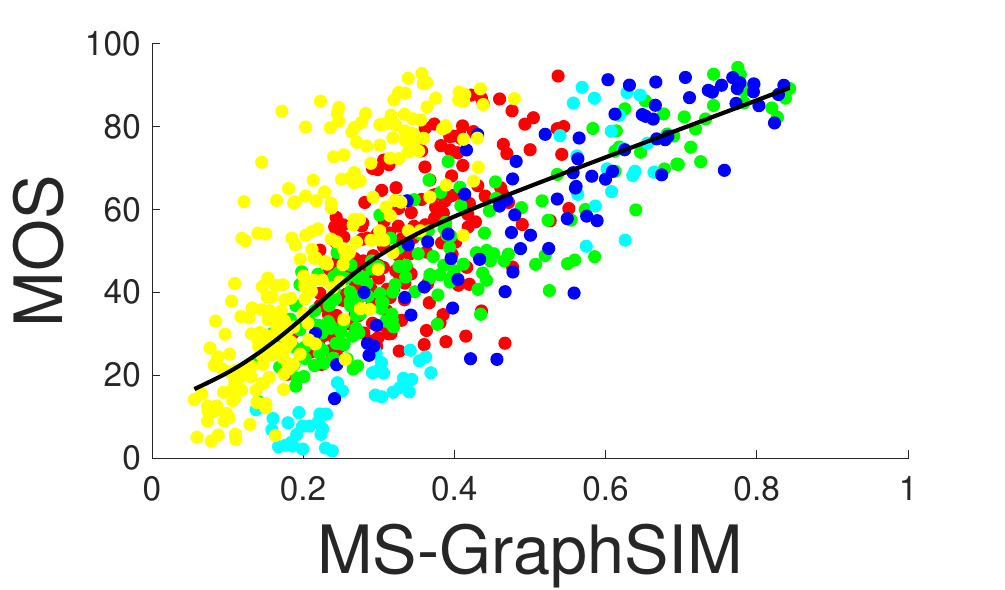}}
    \subfigure[]{\includegraphics[width=0.48\linewidth]{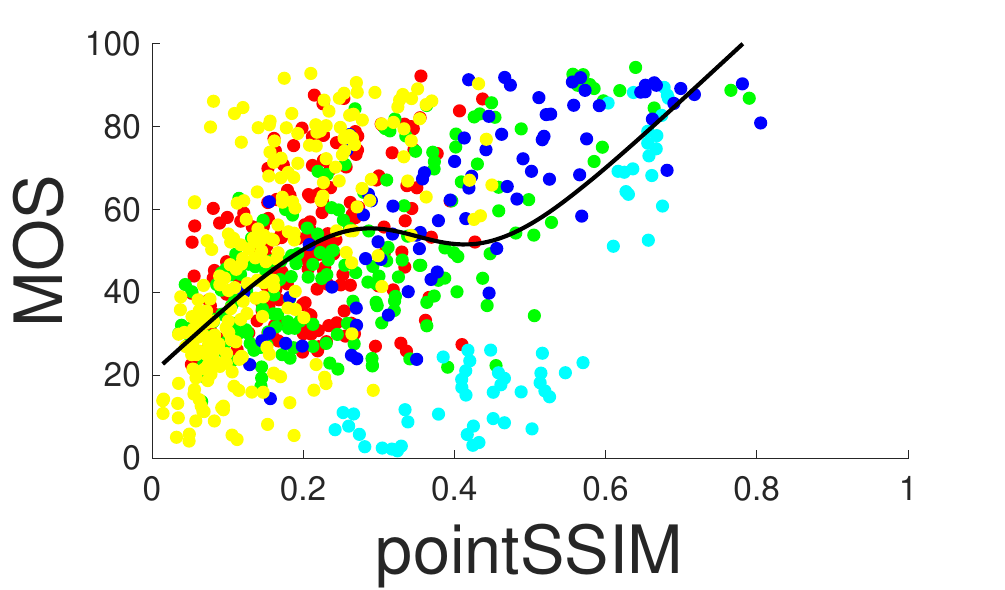}}
    \subfigure[]{\includegraphics[width=0.48\linewidth]{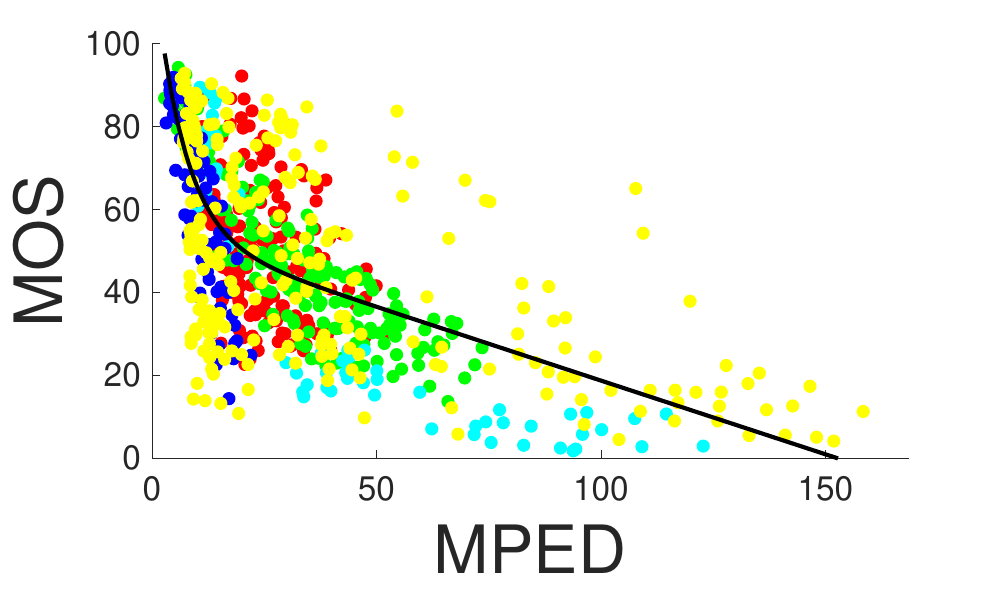}}
    \subfigure[]{\includegraphics[width=0.48\linewidth]{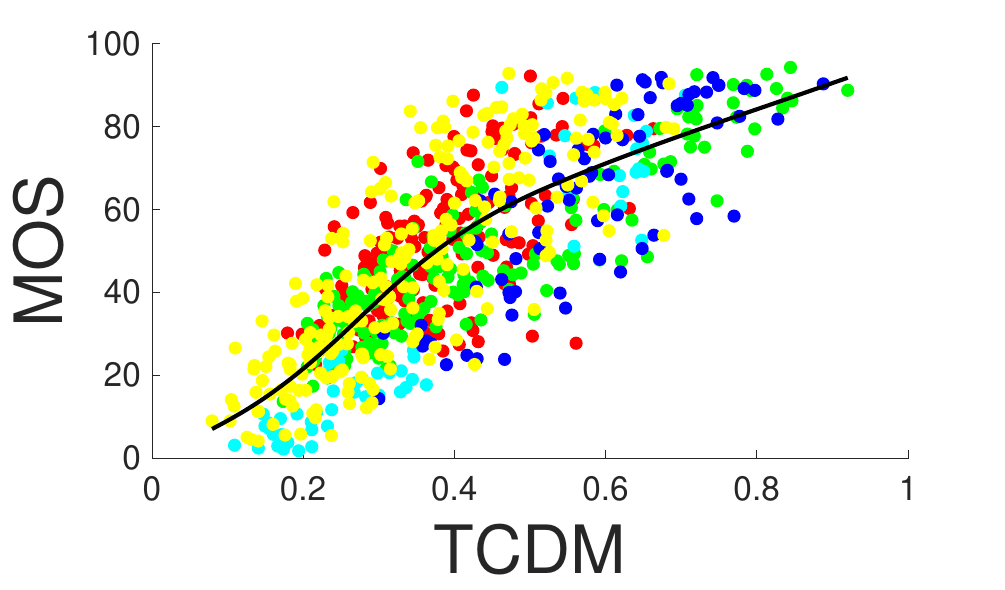}}
    \subfigure[]{\includegraphics[width=0.48\linewidth]{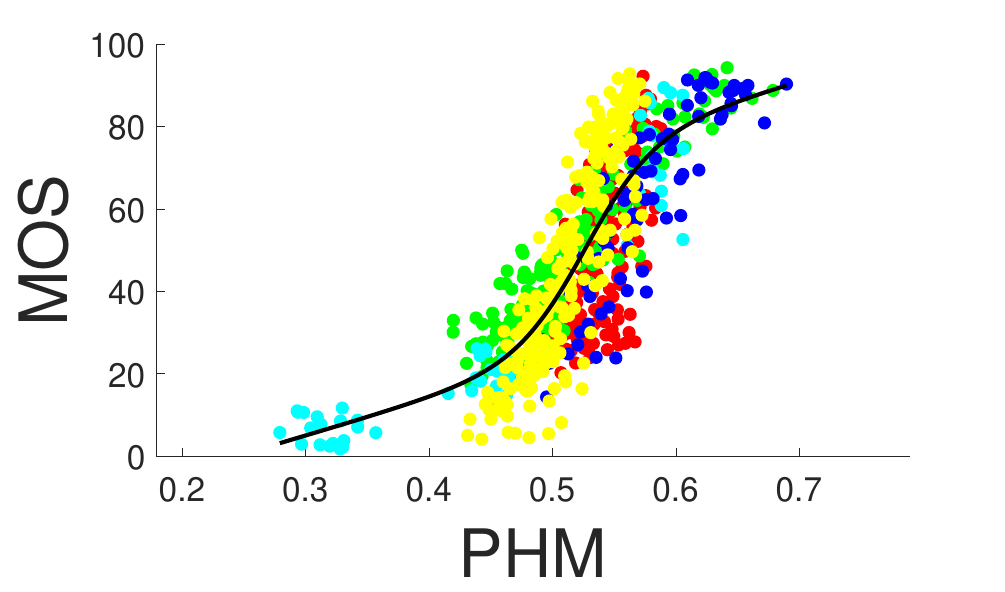}}
    \subfigure{\includegraphics[width=0.8\linewidth]{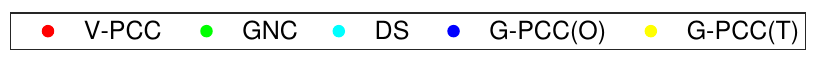}}
    \caption{Scatter plots of subject scores versus objective values obtained by different model prediction on WPC. (a) PCQM. (b) MS-GraphSIM. (c) pointSSIM. (d) MPED. (e) TCDM. (f) PHM.}
    \label{fig:scatter_plot}
    \vspace{-0.6cm}
\end{figure}

\begin{figure*}
    \centering
    \subfigure[]{\includegraphics[width=0.28\linewidth]{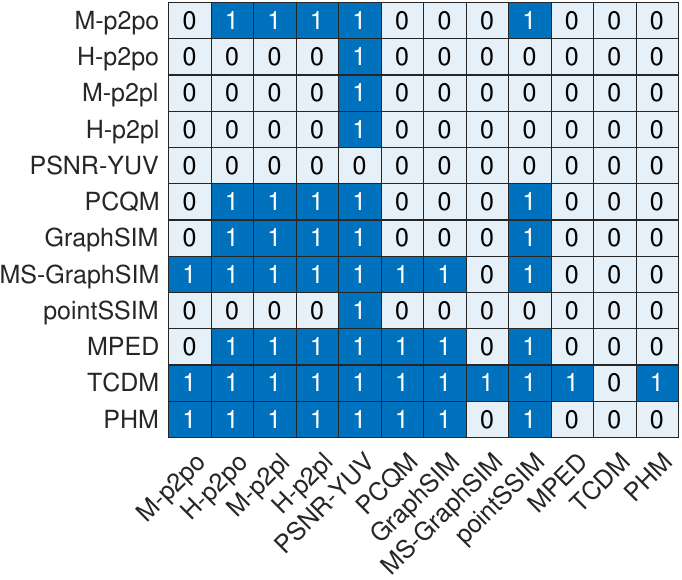}}
    \subfigure[]{\includegraphics[width=0.28\linewidth]{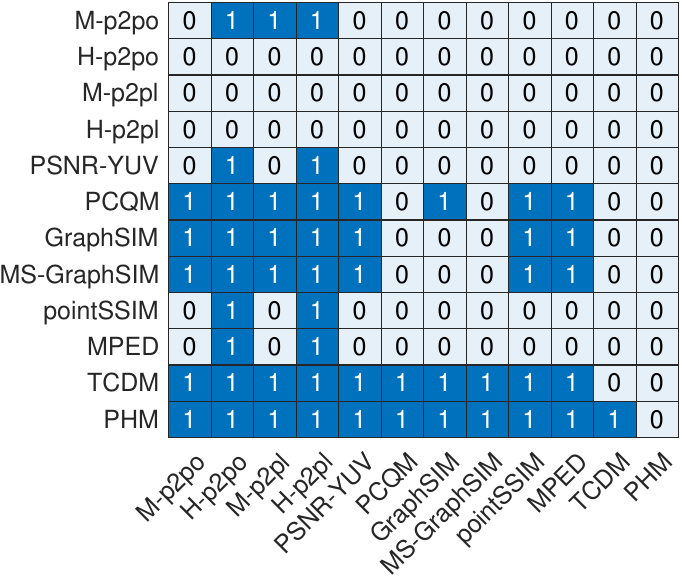}}
    \subfigure[]{\includegraphics[width=0.28\linewidth]{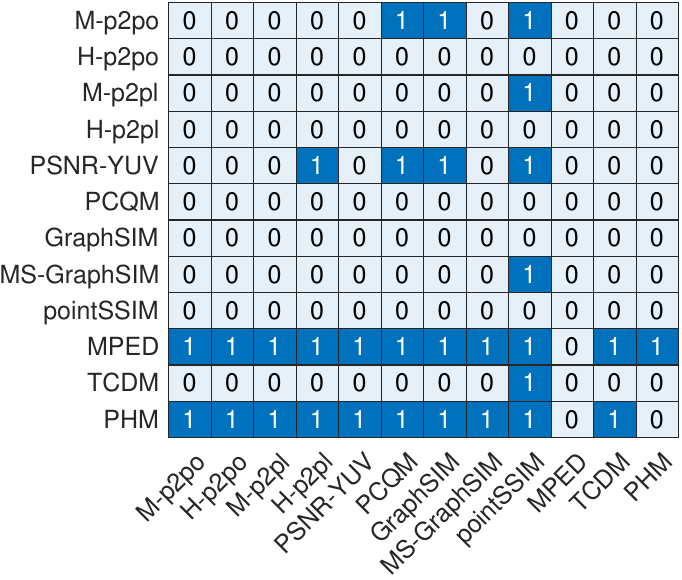}}
    \subfigure[]{\includegraphics[width=0.28\linewidth]{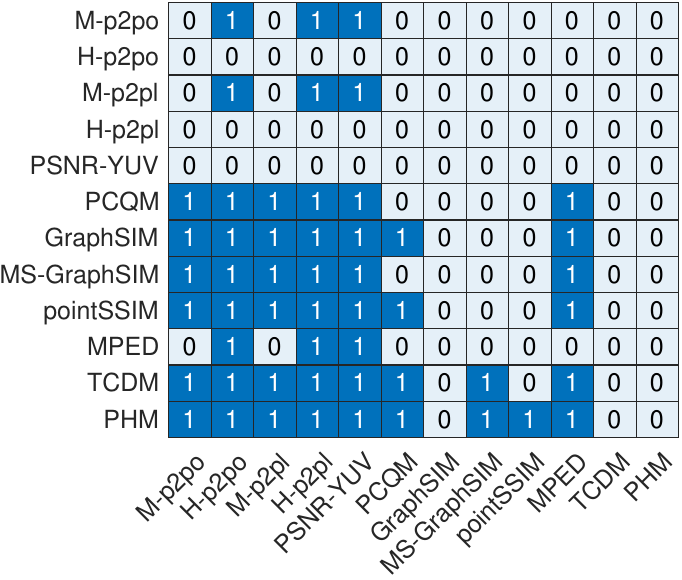}}
    \subfigure[]{\includegraphics[width=0.28\linewidth]{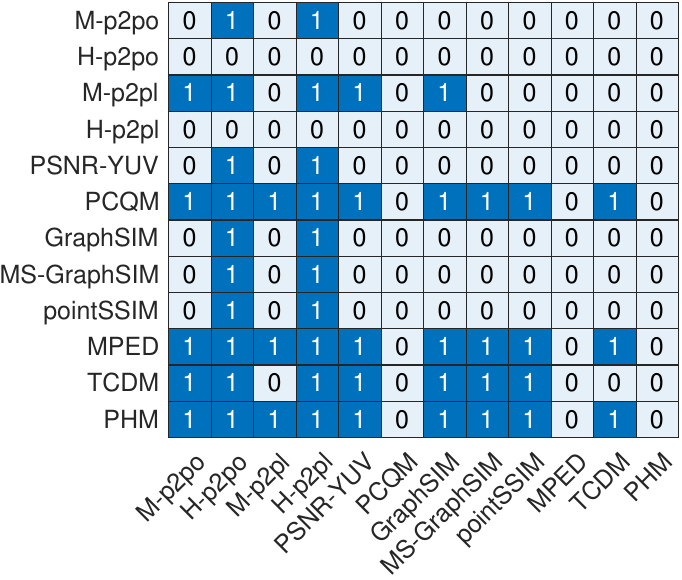}}
    \caption{Illustration of statistical significance tests on the (a) SJTU-PCQA; (b) WPC; (c) LS-PCQA; (d) M-PCCD; (e) ICIP2020.}
    \label{fig:sigficance_test}
    \vspace{-0.4cm}
\end{figure*}

\subsection{Performance Comparison}
Table \ref{tab:overall_performance} lists the experimental results on the five PCQA databases. The proposed metric is compared with 17 FR-PCQA metrics. Note that we use the PSNR measurement for the five point-wise metrics, the L1 norm for MPED \cite{yang2021mped}, and the color measurement for pointSSIM \cite{alexiou2020towards}. For the projection-based metrics, We render  point clouds into six-view images from six perpendicular perspectives (that is, along the positive and negative directions of the x,y,z axes) to facilitate the calculation of projection-based metrics. The quality scores of different views are simply averaged. For each database, the top two results for each evaluation criterion are highlighted in \textbf{boldface}.

From Table \ref{tab:overall_performance},  we can observe that the proposed PHM is among the top two performances in the five databases. In particular, PHM performs well on WPC, MPCCD, and ICIP2020, which shows its sensitivity to compression distortions.
\znote{The projection-based metrics provide relatively poor results. This is mainly because the projection process causes information loss and introduces useless background areas, which impairs the evaluation performance. TCDM provides superior performance than PHM on SJTU-PCQA because it is more sensitive to some geometry distortions (e.g., OT); MPED performs the best on LS-PCQA because it extracts features from relatively small regions (e.g., five nearest neighbors), thus maintaining the sensitivity to microscopic noises that dominate in LS-PCQA. Notably, although some metrics provide outstanding results in certain instances, they perform less well in other cases (e.g., TCDM on LS-PCQA and MPED on M-PCCD). Furthermore, to measure the average performance of these PCQA metrics across multiple databases, we derive the average rank of each metric with respect to competitors in terms of PLCC and SROCC (see the last two columns of Table \ref{tab:overall_performance}). It is evident that PHM outperforms other PCQA metrics with a higher average rank of PLCC and SROCC at (1.6, 1.8), followed by (4.0, 3.8) from TCDM and (4.0, 4.0) from MPED.}

\znote{For better illustration, we provide the scatter plots shown in Fig. \ref{fig:scatter_plot} for multiple competing PCQA metrics (PCQM, MS-GraphSIM, pointSSIM, MPED, TCDM) on the WPC database. In all the plots, each point represents a test point cloud. The black curves in the scatter plots are obtained by the non-linear fitting in Eq. \eqref{eq:logic_fit} 
According to Fig. \ref{fig:scatter_plot}, it can be clearly seen that some metrics usually fail to differentiate the quality among various types of distortion. For example, points representing the "G-PCC(T)" distortion in the MS-GraphSIM plot and points representing the "DS" distortion in the PointSSIM plot significantly deviate from other points, which impairs the overall performance. In comparison, points representing different types of distortion in the PHM plot converge to the fitted curve, which states that PHM can effectively differentiate the quality among various types of distortion and therefore achieves superior overall performance.}

\begin{figure}
    \centering
    \subfigure[] {\includegraphics[width=0.95\linewidth]{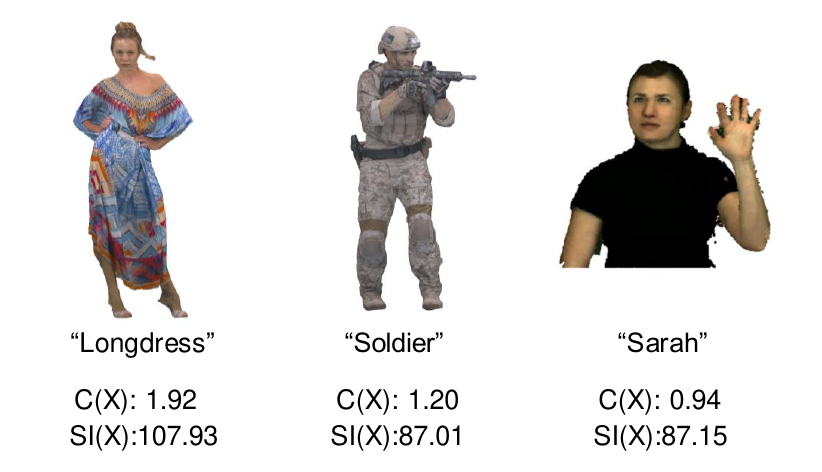}}
    \subfigure[]{\includegraphics[width=0.48\linewidth]{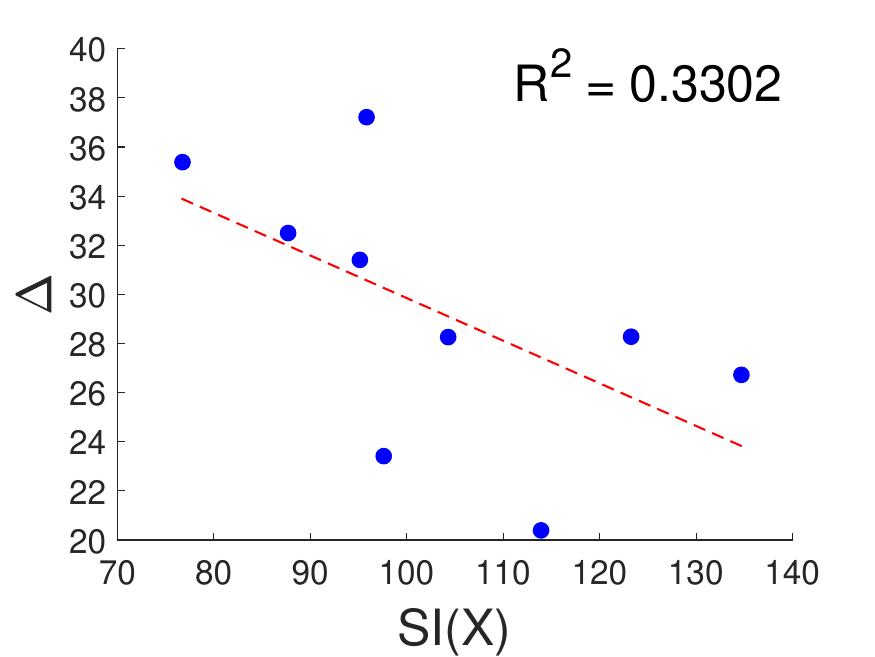}}
    \subfigure[]{\includegraphics[width=0.48\linewidth]{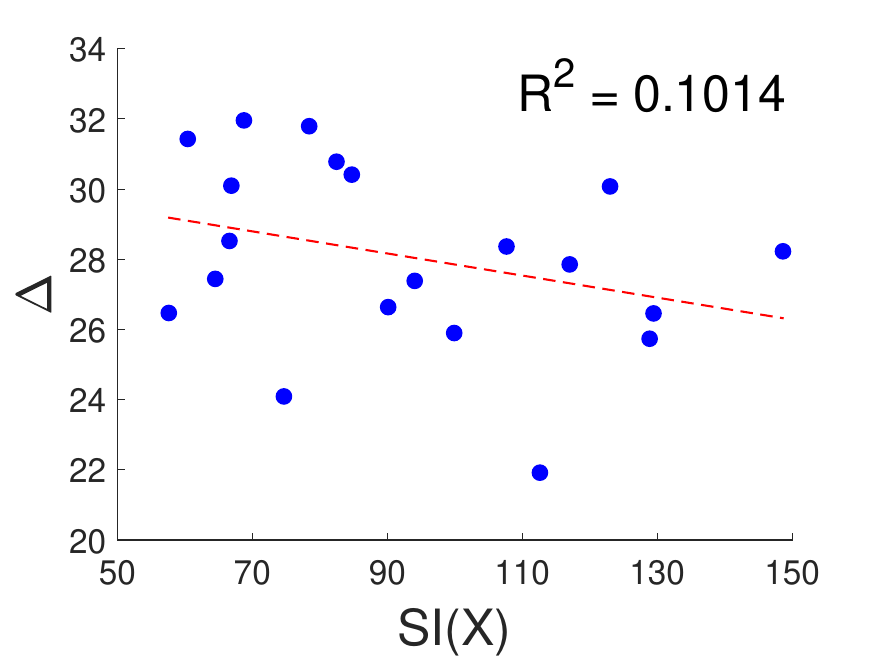}}
    \caption{(a) Exemplified examples of the proposed texture complexity and the spatial information evaluation;
    (b) $SI(\X)-\Delta$ relationship on SJTU-PCQA; (c) $SI(\X)-\Delta$ relationship on WPC.}
    \label{fig:SI_displacement_plot}
    \vspace{-0.4cm}
\end{figure}

We further present the results of statistical significance tests as illustrated in Fig. \ref{fig:sigficance_test}, which is achieved by performing a series of hypothesis tests based on the prediction residuals of each approach after non-linear regression.  We employed the left-tailed F-test to compare the residuals of every pair of models under examination. With a significance level of 0.05, a left-tailed F-test with a value of $H=1$ indicates that the first metric (as denoted by the corresponding row in Fig. \ref{fig:sigficance_test}) exhibits better performance than the second metric (as denoted by the respective column in Fig. \ref{fig:sigficance_test}) with confidence exceeding $95\%$. Conversely, a value of $H=0$ suggests that the first metric does not have a statistically significant advantage over the second metric. Fig. \ref{fig:sigficance_test} (a)–(e) show the significant test results on the five databases, respectively. In total, PHM achieves the value of "1" 47 times, followed by 
TCDM (37 times), MPED (32 times), and PCQM (28 times). This demonstrates that the proposed metric is superior to other metrics. Although TCDM and MPED are competitive with PHM in some specific cases, such as TCDM on SJTU-PCQA and MPED and LS-PCQA. However, this is only valid in a few cases, while PHM performs excellently in the vast majority of instances. From the significance tests, we can find: i) PHM is superior to TCDM on WPC, LS-PCQA, M-PCCD, and ICIP2020. ii) PHM achieves better performance than MPED on SJTU-PCQA, WPC, and M-PCCD. In all, the proposed metric is competitive and promising due to its excellent performance and universality.

\subsection{Effectiveness of the Texture Complexity}

\znote{To further verify the effectiveness of proposed texture complexity, a prevailing complexity index named spatial information (SI) \cite{itu2007subjective} is calculated to replace the texture complexity in the visible difference measurement. More specifically, given the reference point cloud $\X$, we first project it into six views of its bounding box and then obtain the maximum SI value among the six views as $SI(\X)$, following \cite{wu2021subjective}, in which the gradients of the background areas are excluded. We present the texture complexity $C(\X)$ and the spatial information $SI(\X)$ of several samples in Fig. \ref{fig:SI_displacement_plot} (a). From the figure, we can see that the rank of the texture complexity is consistent with human perception. In comparison, the spatial information provides similar values for "Soldier" and "Sarah", which contradicts our intuition. This phenomenon is mainly because the
spatial information actually calculates the standard deviation of image gradients, which is easily influenced by outliers. The "Sarah" point cloud contains an amount of homogeneous regions and obvious edges, and the standard deviation between small gradients (located in homogeneous regions) and large gradients (located in edges) overestimates the complexity of the point cloud.}

\begin{figure}
    \centering
    \includegraphics[width=0.95\linewidth]{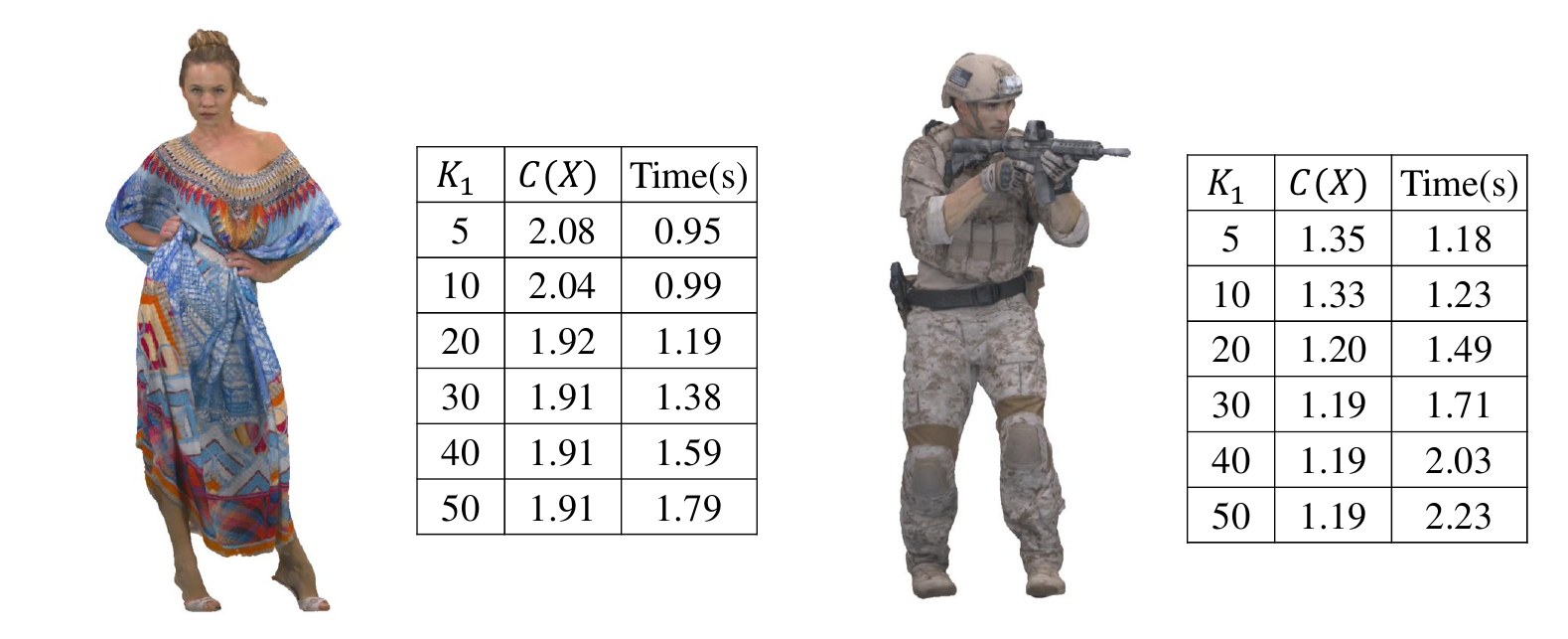}

    \caption{Illustration of the texture complexity values with different $K_1$s.}
    \label{fig:TC_K_trend}
\end{figure}

\begin{table}[t]
  \centering
  \caption{ \MakeUppercase{Performance comparison OF DIFFERENT $K_1$ VAlues on SJTU-PCQA} }
    \begin{tabular}{c|c|c|c}
    \toprule
      {$K_1$} & {PLCC } & {SROCC} & {RMSE} \\
    \midrule
   
5	&0.9054	&0.8855	&1.0301	\\ 
10	&0.9057	&0.8856	&1.0289	\\ 
20	&0.9071	&0.8876	&1.0215	\\ 
30	&0.9069	&0.8876	&1.0225	\\ 
40	&0.9071	&0.8876	&1.0216	\\ 
50	&0.9069	&0.8876	&1.0224	\\ \bottomrule

    \end{tabular}%
  \label{tab:ablation_K1}%

\end{table}%

\znote{Furthermore, we present the relationship between $SI(\X)$ and the horizontal displacement $\Delta$ on SJTU-PCQA and WPC and report the $R^2$ values of linear fitting in Fig. \ref{fig:SI_displacement_plot} (b) and (c). From the figure, we observe that there is neither strong linear relationship nor other proper relationship between the spatial information and displacement, which is in contrast to the high linearity between $C(\X)$ and $\Delta$ in Fig. \ref{fig:complexity_displacement_plot}.
Therefore, we can conclude that the proposed texture complexity is significantly superior to the spatial information in our metric.}

\begin{table}[t]
  \centering
  \caption{ \MakeUppercase{Performance comparison OF DIFFERENT Sample ratios on SJTU-PCQA} }
    \begin{tabular}{c|c|c|c}
    \toprule
      {$L$} & {PLCC } & {SROCC} & {RMSE} \\
    \midrule
   
N/500	&0.9063	&0.8876	&1.0255	\\ 
N/1000	&0.9071	&0.8876	&1.0215	\\ 
N/2000	&0.9074	&\bf 0.8880	&1.0202	\\ 
N/5000  &\bf 0.9076	&0.8877	&\bf 1.0189\\ 
\bottomrule

    \end{tabular}%
  \label{tab:ablation_sampling}%
 
\end{table}%

\begin{table}[t]
  \centering
  \caption{ \MakeUppercase{Performance comparison OF DIFFERENT sub-band numbers on SJTU-PCQA} }
    \begin{tabular}{c|c|c|c}
    \toprule
      {$C$} & {PLCC } & {SROCC} & {RMSE} \\
    \midrule
   
2	&0.9058	&0.8863	&1.0282	\\ 
3	&\bf 0.9071	&\bf 0.8876	&\bf 1.0215	\\ 
4	&0.9050	&0.8859	&1.0322	\\ \bottomrule

    \end{tabular}%
  \label{tab:ablation_subband_num}%
  \vspace{-0.4cm}
\end{table}%

\vspace{-0.6cm}

\subsection{Impact of Parameter Values and Combination Method}
\subsubsection{Impact of the AR order}
\znote{$K_1$ is used as the order of AR model in our method. We determine $K_1=20$ based on multiple factors. First, we illustrate the texture complexity values under different $K_1$ values and the corresponding calculation time in Fig. \ref{fig:TC_K_trend}. From the figure, we have the following observations: i) As $K_1$ increases, the texture complexity first decreases and then stabilizes after $K_1=20$. This is mainly because too small a neighborhood is not enough to describe the center point, while too large a neighborhood results in information redundancy. In fact, only parts of the points in the large neighborhood perform in describing the center point, therefore increasing $K_1$ after reaching the threshold does not significantly increase the self-description ability.
ii) As $K_1$ increases, the calculation time of the texture complexity increases. This is because the time complexity includes $K_1$-related terms, such as $O(K_1^3)$. Therefore, to achieve the balance between self-describing ability and time complexity, we consider $K_1=20$ as a suitable choice.}

\znote{Furthermore, we also conduct experiments to investigate the impact of different $K_1$ values on the overall performance. As illustrated in Table \ref{tab:ablation_K1}, we can see that, with growing $K_1$, the performance first increases and then stabilizes after $K_1=20$. This phenomenon is consistent with the trends of the $C(\X)$ values in Fig. \ref{fig:TC_K_trend} and justifies our parameter selection.}

\begin{table}[H]

  \centering
  \caption{ \MakeUppercase{PERFORMANCE COMPARISON (IN TERMS OF SROCC) OF different combination methods}}
    \resizebox{\linewidth}{!}{
    \begin{tabular}{c|c|c|c|c|c}
    \toprule
    {Combination} & {SJTU-PCQA} & {WPC} & {LS-PCQA} & {M-PCCD} &  {ICIP2020} \\
    \midrule
     $[\M,\M]$ &0.8876	&\textbf{0.8325}	&0.5895	&\textbf{0.9485}	&\textbf{0.9584}
     \\ 
     $[\M,\A]$ &0.8793	&0.8322	&0.5927	&0.9459	&0.9536
     \\ 
     $[\A,\M]$ &\textbf{0.8905}	&0.8258	&0.5909	&0.9467	&0.9583
     \\ 
     $[\A,\A]$ &0.8829	&0.8272	&\textbf{0.5983}	&0.9444	&0.9523
     \\ \bottomrule
    \end{tabular}}
  \label{tab:ablation_combination}%
\end{table}%

\subsubsection{Impact of the Sample Ratio}
\znote{In the appearance degradation measurement, we use the farthest point sampling (FPS) to obtain $L$ seeds that generate the 3D Voronoi diagram, where $L$ determines the number of local patches. We originally follow \cite{yang2020inferring} to set $L=N/1000$. We further test the model performance under different $L$ values and report the results in Table \ref{tab:ablation_sampling}. From the table, we can observe that the performance of PHM is very close for multiple $L$ values, which states that PHM is not very sensitive to the number of patches. }

\begin{table*}[t]

  \centering
  \caption{Ablation study for the key modules. `\Checkmark' or `\XSolidBrush' means the setting is preserved or discarded.}
    \begin{tabular}{c|cc|cc|c|cc|cc}
    \toprule
    \multirow{2}{*}{Index} &\multicolumn{2}{c|}{$D_H$} &\multicolumn{2}{c|}{$D_L$} & \multirow{2}{*}{$\omega$} & \multicolumn{2}{c|}{SJTU} & \multicolumn{2}{c}{WPC}  \\ \cmidrule{2-5} \cmidrule{7-10}
   & $\rm PSNR_Y$ &$C(\X)$ &$D_L^O$ &$D_L^I$ & &PLCC &SROCC &PLCC &SROCC\\
    \midrule
    (1) &\Checkmark  & \XSolidBrush &\XSolidBrush  &\XSolidBrush  &\XSolidBrush &0.7001	&0.6941	&0.6174	&0.5955

 \\
    (2) &\Checkmark  & \Checkmark &\XSolidBrush  &\XSolidBrush  &\XSolidBrush &0.8185	&0.8199	&0.7004	&0.6762

    \\ \midrule
    (3) &\XSolidBrush  & \XSolidBrush & \Checkmark  &\XSolidBrush  &\XSolidBrush &0.7536	&0.6912	&0.5221	&0.4542

  \\
    (4) &\XSolidBrush  & \XSolidBrush & \XSolidBrush  &\Checkmark  &\XSolidBrush &0.7321	&0.7296	&0.5974	&0.5546

   \\
    (5) &\XSolidBrush  & \XSolidBrush & \Checkmark   &\Checkmark  &\XSolidBrush &0.7992	&0.7855	&0.7308	&0.7161

    \\ \midrule
    (6)&\Checkmark   &\Checkmark & \Checkmark   &\Checkmark  &\XSolidBrush &0.8881	&0.8712	&0.8333	&0.8269

    \\
   (7) &\Checkmark   &\Checkmark & \Checkmark   &\Checkmark  &\Checkmark &\textbf{0.9071}	&\textbf{0.8876}	&\textbf{0.8391}	&\textbf{0.8325}

    \\
    \bottomrule
    \end{tabular}%
  \label{tab:ablation_study}%
   
\end{table*}%

\subsubsection{Impact of the Sub-band Number}

\znote{We utilize the SGWT to decompose the luminance channel of each patch into one low-pass sub-band and $C=3$ band-pass sub-bands. In this part, we test the model performance with different $C$ values and show the results in Table \ref{tab:ablation_subband_num}. We see that $C=3$ provides the best performance in terms of the three criteria. In general, too small $C$ may not be enough to distinguish the characteristics of different frequency bands, while too large $C$ can result in information redundancy. Therefore, $C=3$ is a suitable choice for our metric.}

\subsubsection{Impact of the Combination Method}

\znote{In this part, we perform the performance comparison of different combination methods. Concretely, we separately adopt \textit{multiplication} (denoted by $\M$) and  \textit{averaging} (denoted by $\A$) for $(D_L^O,D_L^I)$ and $(D_H,D_L)$. The specific combination forms are shown as follows: }
\begin{equation}
\begin{aligned}
    \M(D_L^O,D_L^I)&=\sqrt{D_L^O\cdot D_L^I},\\
    \A(D_L^O,D_L^I)&=(D_L^O+ D_L^I)/2,\\
    \M(D_H,D_L)&=D_H^{1-\omega}\cdot D_L^{\omega},\\
    \A(D_H,D_L)&=(D_H^{1-\omega}+ D_L^{\omega})/2.
\end{aligned}\nonumber
\end{equation}
\znote{Based on the above equation, four different combination lists can be derived and the results are summarized in Table \ref{tab:ablation_combination}. From the table, we can see that none of these combination lists produce severe performance degradation. The combination list used in PHM, that is, $[\M,\M]$, achieves the best performance on three databases, which makes it a suitable choice for our metric.}

\subsection{Ablation Study}
\znote{There are several components involved in our model, i.e., the visible difference measurement $D_H$ that compensates $\rm PSNR_Y$ using the texture complexity $C(\X)$, the appearance degradation measurement $D_L$ that combines geometry and attribute component $D_L^0$ and $D_L^I$, and the adaptive combination weight $\omega$. To investigate the individual contribution of these components, we perform the ablation study for the components adopted in the hybrid metric and report the results in Table \ref{tab:ablation_study}. \zznote{In the table, index (1) represents only using $\rm PSNR_Y$ for quality prediction, while (2) represents the full $D_H$ that compensates $\rm PSNR_Y$ with $C(\X)$; (3) and (4) respectively indicate  leveraging individual $D_L^O$ or $D_L^I$, while (5) denotes the full $D_L$ that integrates $D_L^O$ and $D_L^I$; (6) performs direct multiplication for two measurements (i.e., $D_H\cdot D_L$), while (7) denotes the full metric using the non-linear combination manner (i.e., $D_H^{1-\omega}\cdot D_L^\omega$).}
The following observations can be obtained from the table: i) According to (1) and (2), we can see that introducing the texture complexity significantly improves the performance of $D_H$. ii) From (3)-(5), it is observed that both $D_L^O$ and $D_L^I$ contribute to the evaluation of appearance degradation. iii) Seeing (6) and (7), we can conclude that combining the two components in a non-linear manner  achieves significant gain compared to direct multiplication. This demonstrates that the adaptive combination is effective to model human perception in the hybrid metric. }

\section{Conclusion}\label{sec:conclusion}
In this paper, based on the assumption that HVS flexibly handles visual information for different distortion levels, we have proposed a novel FR-PCQA metric that dynamically combines visible difference and appearance degradation. By merging the characteristics of the HVS into the measurements of
the two components and adopting an adaptive combination manner, the proposed metric is able to assess point cloud quality more accurately. Experimental results on various databases demonstrate that PHM achieves state-of-the-art performance.

\ifCLASSOPTIONcaptionsoff
  \newpage
\fi

\bibliographystyle{IEEEtran}
\bibliography{ref}

\newpage
\appendices

\begin{table*}[t]
  \centering
  \caption{PERFORMANCE COMPARISON of FR-PCQA METRICS ON EACH INDIVIDUAL DISTORTION TYPE IN TERMS OF SROCC}
\resizebox{\linewidth}{!}{
\begin{tabular}{c|ccccccccc|c}
     \hline
    SJTU-PCQA   & {M-p2po}  & {M-p2pl} & {$\rm PSNR_{YUV}$} & {PCQM} & {GraphSIM} & {MS-GraphSIM} & {PointSSIM} & {MPED} & {TCDM} & Proposed \\
    \hline
     OT &\bf{0.825}	&\bf{0.849}		&0.357	&0.758	&0.693	&0.714	&0.756 &0.679	&0.793 & 0.716 \\
   
     CN &-	&-	&0.753	&\bf{0.842}	&0.778	&0.770	&0.797 &{0.824}	&{0.819} & \bf{0.848}
     \\
   
    DS   &0.812	&0.478	&0.542	&0.808	&{0.872}	&0.864	&0.816 &\bf{0.878}	&{0.876} & \bf{0.889}\\

    GGN  &\bf{0.950}		&\bf{0.936}	&0.675	&0.905	&0.916	&0.916	&0.916 &0.925	&0.921 & 0.924
     \\
    
    D+C  &0.885	&0.566	&0.862	&{0.922}	&0.886	&0.914	&0.842 &\bf{0.937}	&\bf{0.934} & {0.931}
    \\
    
    D+G   &0.934 &0.926	&0.61	&0.882	&0.888	&0.905	&0.913 &0.928	&\bf{0.944} &\bf{0.937}
    \\
   
    C+G &0.951	&0.950	&0.852	&0.922	&0.941	&0.951	&0.851 &\bf{0.969}	&\bf{0.951}  &{0.949}\\

    \hline
       \multicolumn{1}{r}{} & \multicolumn{1}{r}{} & \multicolumn{1}{r}{} & \multicolumn{1}{r}{} & \multicolumn{1}{r}{} & \multicolumn{1}{r}{} & \multicolumn{1}{r}{} & \multicolumn{1}{r}{} & \multicolumn{1}{r}{} & \multicolumn{1}{r}{} &   \\
    
     \hline
    WPC   & {M-p2po} & {M-p2pl} & {$\rm PSNR_{YUV}$} & {PCQM} & {GraphSIM} & {MS-GraphSIM} & {PointSSIM} &MPED &{TCDM} & Proposed\\
    \hline
    DS &\bf{0.901}	&0.849	&0.707	&0.875	&\bf{0.898}	&0.887	&0.836 &0.897	&0.882 & 0.884\\

    GNC &0.729	&0.738	&0.777	&\bf{0.886}	&0.840	&{0.869}	&0.587 &{0.880}	&{0.857} &\bf{0.894} \\
    
    G-PCC(O)  &-	&-		&0.826	&\bf{0.894}	&{0.855}	&{0.859}	&0.792	&{0.869}  &0.795 &\bf{0.882}\\
   
    G-PCC(T) &0.465	&0.463 &0.646	&0.821	&0.816	&\bf{0.839}	&0.681 &0.551	&{0.832} &\bf{0.877}\\
  
    V-PCC  &\bf{0.698}	&\bf{0.705}	&0.345	&0.643	&0.612	&0.631	&0.366 &0.476	&0.640 &0.591\\
    \hline

    \multicolumn{1}{r}{} & \multicolumn{1}{r}{} & \multicolumn{1}{r}{} & \multicolumn{1}{r}{} & \multicolumn{1}{r}{} & \multicolumn{1}{r}{} & \multicolumn{1}{r}{} & \multicolumn{1}{r}{} & \multicolumn{1}{r}{} & \multicolumn{1}{r}{} & \\
    
     \hline
    M-PCCD   & {M-p2po} & {M-p2pl} & {$\rm PSNR_{YUV}$} & {PCQM} & {GraphSIM} & {MS-GraphSIM} & {PointSSIM} &MPED &{TCDM} & Proposed\\
    \hline
    G-PCC(O)-lifting	&0.900	&0.899	&0.722	&0.932	&0.963	&{0.947}	&0.940 &0.904	&\bf{0.975}	&\bf{0.967}\\
    G-PCC(O)-RAHT	&0.884	&0.879 &0.735	&0.923	&{0.952}	&0.939	&0.940 &0.901 &\bf{0.969} &\bf{0.959}\\
    G-PCC(T)-lifting&0.761	&0.822	&0.659	&0.893	&\bf{0.938}	&{0.917}	&{0.900} &0.834	&0.898	&\bf{0.957}\\
    G-PCC(T)-RAHT	&0.708	&0.767	&0.622	&0.895	&{0.927}	&0.898	&\bf{0.958} &0.798	&0.899	&\bf{0.962}\\
    VPCC	&0.277	&0.550 &0.317	&0.736	&\bf{0.869}	&0.824	&{0.845} &0.531	&0.814	&\bf{0.857}\\

    \hline
    
    \multicolumn{1}{r}{} & \multicolumn{1}{r}{} & \multicolumn{1}{r}{} & \multicolumn{1}{r}{} & \multicolumn{1}{r}{} & \multicolumn{1}{r}{} & \multicolumn{1}{r}{} & \multicolumn{1}{r}{} & \multicolumn{1}{r}{} & \multicolumn{1}{r}{} & \\
    
     \hline
    ICIP2020   & {M-p2po} & {M-p2pl}  & {$\rm PSNR_{YUV}$} & {PCQM} & {GraphSIM} & {MS-GraphSIM} & {PointSSIM} &MPED &{TCDM} & Proposed \\
    \hline
    GPCC(O)	&0.904	&\bf{0.942}	&0.885	&\bf{0.968}	&0.831	&0.855	&0.812 &0.899	&0.885 &{0.939}\\
    GPCC(T)	&0.908	&0.892	&0.817	&{0.954}	&0.941	&0.939	&0.910 &\bf{0.955}	&\bf{0.970} &{0.947}	\\
    VPCC	&0.723	&0.828	&{0.848} &\bf{0.960}	&0.720	&0.770	&0.822 &{0.904}	&0.822	&\bf{0.952}\\

    \hline
    
    \multicolumn{1}{r}{} & \multicolumn{1}{r}{} & \multicolumn{1}{r}{} & \multicolumn{1}{r}{} & \multicolumn{1}{r}{} & \multicolumn{1}{r}{} & \multicolumn{1}{r}{} & \multicolumn{1}{r}{} & \multicolumn{1}{r}{} & \multicolumn{1}{r}{} & \\
    
     \hline
    LS-PCQA   & {M-p2po} & {M-p2pl}  & {$\rm PSNR_{YUV}$} & {PCQM} & {GraphSIM} & {MS-GraphSIM} & {PointSSIM} &MPED &{TCDM} & Proposed \\
    \hline
 Color Noise	&-	&-	&0.834	&0.815	&0.687	&0.715	&\bf{0.863}	&0.839	&0.586	&\bf{0.876}	\\
Color Quantization Dither	&-	&-	&\bf{0.826}	&\bf{0.769}	&0.517	&0.551	&0.384	&0.759	&0.598	&0.521	\\
Contrast Distortion	&-	&-	&0.689	&\bf{0.744}	&0.679	&0.815	&0.542	&0.682	&\bf{0.886}	&0.579	\\
Correlated Gaussian Noise	&-	&-	&\bf{0.939}	&0.851	&0.589	&0.643	&0.572	&\bf{0.875}	&0.579	&0.628	\\
Down-sampling	&\bf{0.881}	&0.627	&0.015	&0.525	&\bf{0.843}	&0.688	&0.342	&0.441	&0.554	&0.672	\\
Gamma Noise	&-	&-	&0.750	&0.708	&0.588	&0.437	&0.396	&\bf{0.882}	&
\bf{0.765}	&{0.669}	\\
Gaussian Noise	&-	&-	&0.645	&0.770	&0.483	&0.772	&\bf{0.84}	&0.583	&0.437	&\bf{0.877}	\\
Gaussian Shifting	&0.741	&0.719	&0.755	&\bf{0.816}	&0.743	&0.741	&0.504	&\bf{0.786}	&0.665	&{0.694}	\\
High-frequency Noise	&-	&-	&0.836	&\bf{0.915}	&0.763	&0.590	&0.715	&\bf{0.890}	&0.691	&{0.825}	\\

Local Loss	&0.536	&0.498	&0.689	&0.771	&\bf{0.871}	&\bf{0.861}	&0.852	&0.693	&0.854	&0.829	\\
Local Offset	&\bf{0.937}	&0.935	&0.667	&0.852	&0.906	&0.906	&\bf{0.95}	&0.932	&0.926	&0.573	\\
Local Rotation	&\bf{0.82}	&0.713	&0.327	&0.657	&0.724	&0.711	&0\bf{.801}	&0.689	&0.787	&0.631	\\
Luma Noise	&-	&-	&0.773	&0.748	&\bf{0.818}	&\bf{0.813}	&0.542	&0.794	&{0.811}	&0.762	\\
Mean Shift	&-	&-	&0.422	&0.615	&0.706	&{0.754}	&0.617	&\bf{0.763}	&0.581	&\bf{0.797}	\\
Multiplicative Gaussian Noise	&-	&-	&0.751	&{0.754}	&0.648	&0.627	&\bf{0.756}	&\bf{0.78}	&0.597	&0.674	\\
Poisson Noise	&-	&-	&\bf{0.682}	&\bf{0.663}	&0.422	&0.489	&0.056	&0.475	&0.088	&0.167	\\
Quantization Noise	&-	&-	&0.781	&\bf{0.848}	&0.618	&0.497	&0.672	&\bf{0.853}	&0.455	&0.785\\
Rayleigh Noise	&-	&-	&\bf{0.894}	&\bf{0.838}	&0.707	&0.724	&0.487	&0.802	&0.585	&0.539\\
Saltpepper Noise	&-	&-	&0.395	&0.638	&0.560	&0.516	&\bf{0.725}	&\bf{0.719}	&0.541	&0.631	\\
Saturation Distortion	&-	&-	&0.739	&\bf{0.851}	&0.703	&0.759	&0.446	&0.718	&\bf{0.824}	&0.573	\\
Uniform Noise	&-	&-	&\bf{0.898}	&0.686	&0.646	&0.665	&0.504	&\bf{0.775}	&0.564	&0.568	\\
Uniform Shifting	&\bf{0.852}	&\bf{0.852}	&0.797	&0.639	&\bf{0.870}	&0.809	&0.829	&0.748	&0.751	&0.745	\\
V-PCC &0.753	&0.753	&0.372	&0.796	&0.808	&\bf{0.865}	&0.456	&\bf{0.901}	&0.684	&0.738\\
AVS-PCC(limitlossyG-lossyA)	&\bf{0.935}	&\bf{0.946}	&0.882	&-	&0.744	&0.700	&0.712	&0.917	&0.900	&{0.933}	\\
AVS-PCC(losslessG-limitlossyA)	&-	&-	&0.838	&-	&\bf{0.877}	&\bf{0.877}	&0.832	&0.864	&0.776	&0.849	\\
AVS-PCC(losslessG-lossyA)	&-	&-	&\bf{0.916}	&-	&0.796	&0.825	&0.890	&0.893	&0.808	&\bf{0.908}	\\
G-PCC(losslessG-lossyA)	&-	&-	&0.568	&\bf{0.862}	&0.665	&0.550	&0.624	&\bf{0.878}	&0.806	&0.677	\\
G-PCC(losslessG- nearlosslessA)	&-	&-	&0.878	&\bf{0.938}	&0.899	&0.885	&0.876	&0.909	&0.804	&\bf{0.933}	\\
G-PCC(lossyG-lossyA)	&\bf{0.955}	&0.926	&0.731	&0.845	&0.867	&0.898	&0.465	&\bf{0.941}	&0.859	&0.829	\\
Octree	&{0.779}	&\bf{0.788}	&0.524	&0.676	&0.758	&0.758	&0.045	&\bf{0.836}	&0.660	&0.727	\\
Possion Reconstruction	&\bf{0.847}	&\bf{0.835}	&0.213	&0.721	&0.648	&0.772	&0.313	&0.308	&0.652	&0.376	\\

    \hline
    \end{tabular}}\label{tab:test}
\end{table*}

\section{Additional Experiment} \label{FirstAppendix}
\subsection{Performance Comparison on Individual Distortion Type}

To more comprehensively evaluate an PCQA metric’s ability to predict point cloud degradation caused by specific types of distortion, we compare the performance of competing methods on each type of distortion. The results in terms of SROCC are illustrated in Table \ref{tab:test}.
There are a total of 51 groups of distorted samples in the five databases and we highlight the top two results for each group using \textbf{boldface}. SROCC is chosen because it is suitable for measuring a small number of data points and its value will not be affected by an unsuccessful monotonic nonlinear mapping.
\begin{figure}[t]
    \centering
        \includegraphics[width=\linewidth]{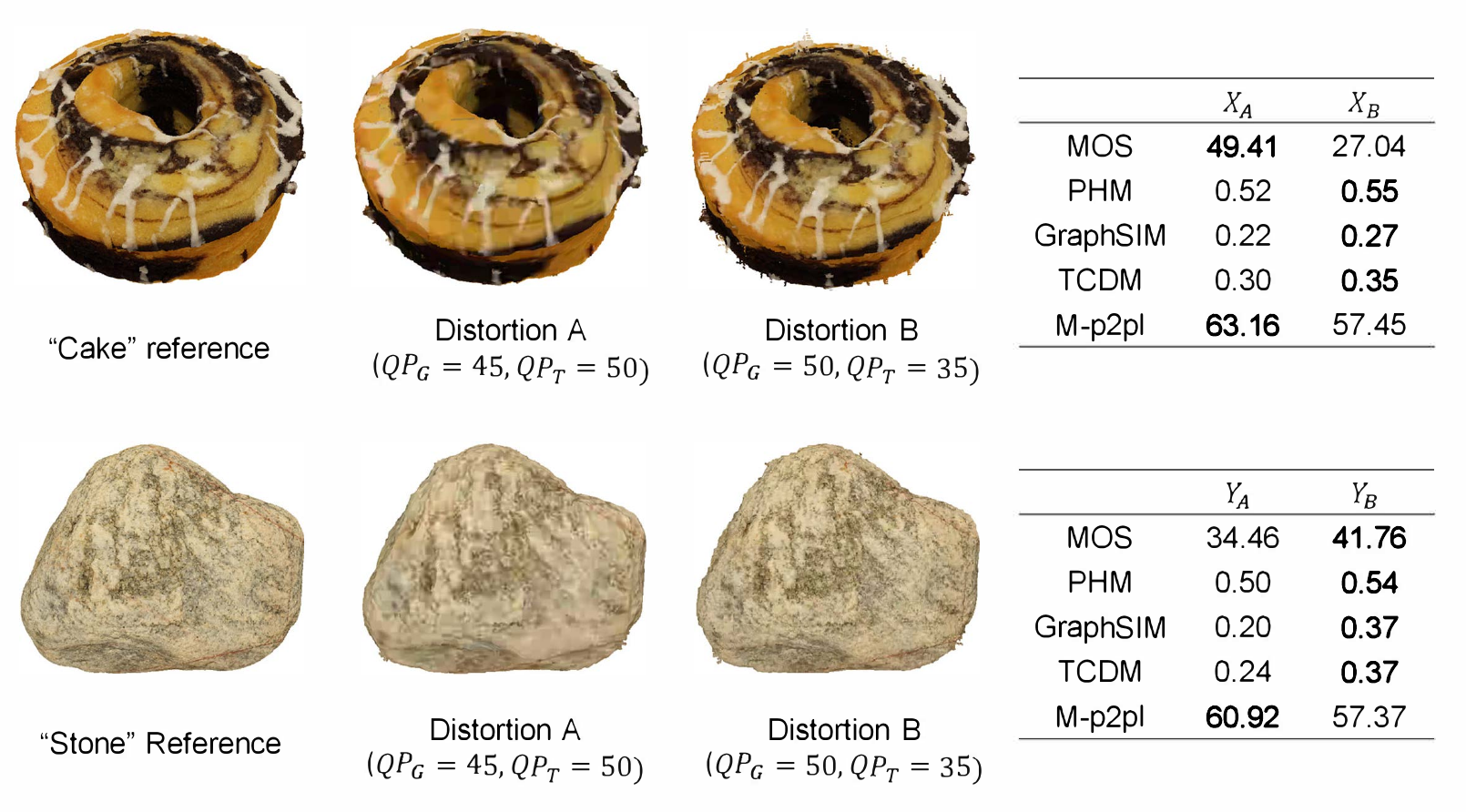}

    \caption{Example point clouds corrupted by V-PCC on the WPC database. $QP_G$ and $QP_T$ represent the geometry and texture quantization parameters, respectively. The table on the right side illustrate the predictions of objective metrics and predictions corresponding to higher quality are highlighted in \textbf{boldface}. }
    \label{fig:WPC_VPCC_sample}
\end{figure}

From Table \ref{tab:test}, we can see that PHM is among the top two metrics 17 times, followed by MPED and PCQM, which are among the top two models 17 times and 15 times, respectively. More specifically, PHM is superior to other metrics on SJTU-PCQA, WPC, M-PCCD and ICIP2020, ranking among the top two for 12 times in 20 groups, followed by PCQM (5 times) and TCDM (4 times). Note that WPC, M-PCCD, and ICIP2020 mostly contain a variety of compression distortions. Therefore, the promising results of PHM on these databases demonstrate its effectiveness when encountering most compression modes, making it a potential indicator to guide the optimization of the emerging point cloud compression research.

\begin{table}[t]
  \centering
  \caption{ \MakeUppercase{PERFORMANCE COMPARISON (IN TERMS OF SROCC) OF DIFFERENT Wavelet functions}}
    \resizebox{\linewidth}{!}{
    \begin{tabular}{c|c|c|c|c|c}
    \hline
    {Wavelet type} & {SJTU-PCQA} & {WPC} & {LS-PCQA} & {M-PCCD} &  {ICIP2020} \\
    \hline
     Mexican hat \cite{perraudin2016gspbox} &0.8826	&\bf0.8354	&0.5957	&\bf 0.9499	&\bf0.9626
 \\ 
    \hline
     Meyer \cite{perraudin2016gspbox} &\bf 0.8906	&0.8237	&\bf0.5999	&0.9470	&0.9569 \\ \hline
     Ref. \cite{hammond2011wavelets} &0.8876	&0.8325	&0.5895	&0.9485	&0.9584
 \\ \hline
    \end{tabular}}
  \label{tab:wavelet_type}%
\end{table}%

It is noticeable that PHM provides relatively poor performance for the V-PCC distortion on WPC. Further looking into this issue, we found that PHM usually fails in some cases. Referring to Fig. \ref{fig:WPC_VPCC_sample}, distortions A and B refer to two different groups of the V-PCC quantification parameter (QP). The distortion A is composed of a lower geometry QP (denoted by $QP_G$) and a larger texture QP (denoted by $QP_T$) while the distortion B is the opposite. Given two reference point clouds, we denote their distorted versions affected by the two groups of QP as $(X_A,X_B)$ and $(Y_A,Y_B)$. From the figure, we can see that PHM tends to present higher scores for distortion B regardless of the corresponding references. However, human observers provide a higher MOS for $X_A$ than $X_B$, but a lower score for $Y_A$ than $Y_B$. This is mainly because $X_B$ has more apparent geometry degradation than $Y_B$. In fact, human subjects place higher priority on geometry degradation than texture degradation, leading to opposite trends for the two groups of samples. The existing methods, including the proposed PHM, fail to exploit such a visual property. One possible solution is to apply the adaptive mechanism for the combination of geometry and texture features in quality metrics. For example, a higher weight could be assigned to the geometry features when the geometry degradation is severe.

LS-PCQA contains 31 types of distortions, including 16 types of color distortions that do not distort any geometry information, 8 types of compression distortions, and 7 types of other distortions. The proposed method becomes the top two metrics 5 times, demonstrating inferior performance compared to MPED (13 times). MPED presents favorable performance for color distortions because it extracts features from relatively small regions (e.g., five nearest neighbors), thus maintaining superior sensitivity to these texture noises.

\begin{figure}[t]
    \centering
    \subfigure[]{\includegraphics[width=\linewidth]{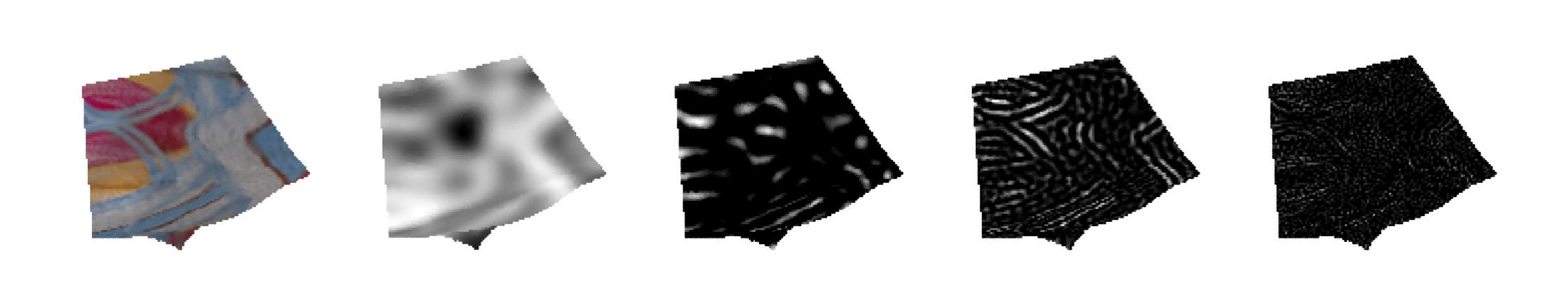}}
    \subfigure[]{\includegraphics[width=\linewidth]{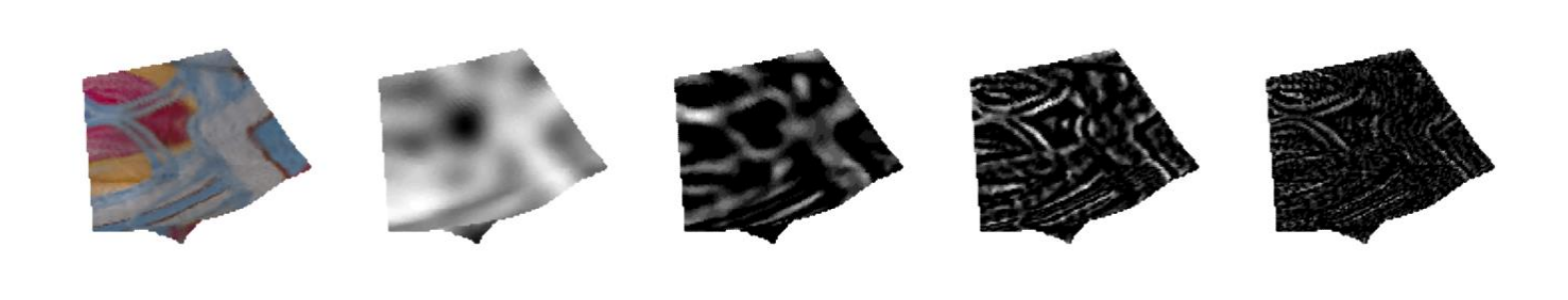}}
    \subfigure[]{\includegraphics[width=\linewidth]{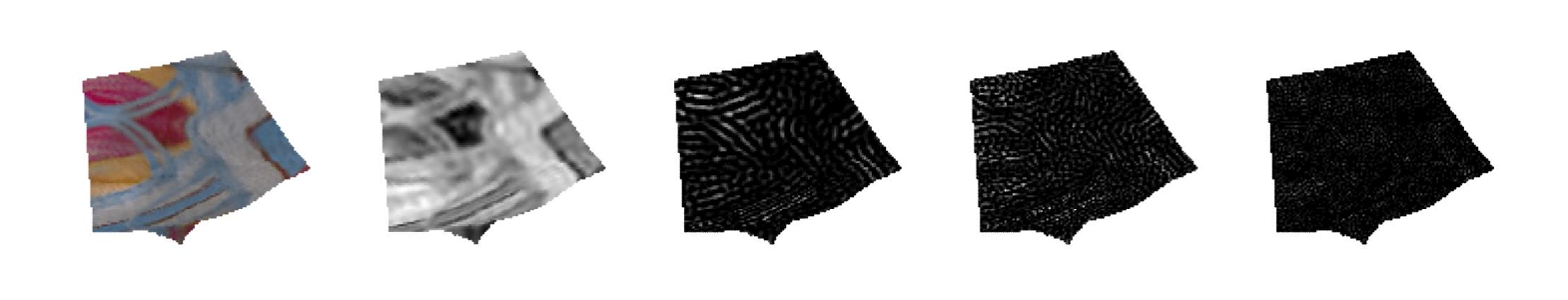}}
    \caption{Illustration of sub-bands obtained from different types of wavelets. (a) Wavelet proposed in \cite{hammond2011wavelets}. (b) Mexican hat wavelet provided by \cite{perraudin2016gspbox}. (c) Meyer wavelet provided by \cite{perraudin2016gspbox}.}
    \label{fig:wavelet_type}
\end{figure}

\subsection{Impact of Wavelet Type}
In our paper we choose the graph wavelet proposed in \cite{hammond2011wavelets} to achieve the multi-resolution decomposition of the point cloud texture. To study the performance impact of different wavelet types,  we conduct additional experiments using two other wavelet types: Mexican hat wavelet and Meyer wavelet provided by \cite{perraudin2016gspbox}.  We depict the sub-bands obtained from the three wavelet types in Fig. \ref{fig:wavelet_type} to illustrate the differences among them. From Fig. \ref{fig:wavelet_type}, it is evident that the $0$-th sub-band decomposed by the Meyer wavelet (Fig. \ref{fig:wavelet_type} (c))) preserves more fine-grained information, demonstrating that it retains a higher amount of energy in the low-frequency sub-band. In comparison, the Mexican hat wavelet preserves more energy in high-frequency sub-bands, presenting more information in the $3$-th sub-bands. The above observations demonstrate that different types of wavelets can generate various decomposition results.

Subsequently, we evaluate the performance of the three wavelets on the five databases and report the experimental results in Table \ref{tab:wavelet_type}. From the results, we can see that the Mexican hat wavelet outperforms \cite{hammond2011wavelets} on SJUT-PCQA and LS-PCQA, and the Meyer wavelet provides better performance on WPC, M-PCCD, and ICIP2020. In all, all wavelets show competitive performance compared to most existing quality metrics, highlighting the effectiveness and universality of the SGWT.

\begin{table}
  \centering
  \caption{ \MakeUppercase{PERFORMANCE COMPARISON (IN TERMS OF SROCC) OF DIFFERENT $N_b$}}
    \resizebox{\linewidth}{!}{
    \begin{tabular}{c|c|c|c|c|c}
    \hline
    {$N_b$} & {SJTU-PCQA} & {WPC} & {LS-PCQA} & {M-PCCD} &  {ICIP2020} \\
    \hline
     10 &\bf0.890	&0.806	&0.594	&0.942	&0.957
 \\ 
    \hline
     30 &0.888	&0.831	&0.588	&0.948	&0.958

     \\ \hline
     50 &0.888	&\bf0.833	&0.589 &\bf0.948	&0.958

     \\ \hline
     70 &0.888	&0.829	&0.597	&0.946	&\bf0.959
     \\ \hline
     100 &0.888	&0.815	&\bf0.607	&0.941	&0.958

     \\ \hline
    \end{tabular}}
  \label{tab:bin_num}%
\end{table}%

\subsection{Impact of WCM Binning}
The number of bins influences the precision of the WCM. We further test the model performance (in terms of SROCC) with different $N_b$ and show the results in Table \ref{tab:bin_num}. We can see that the determined parameter $N_b=50$ presents the best performance on WPC and M-PCCD. Meanwhile, it is observed that too low $N_b$ weakens the performance slightly on some databases (e.g., WPC). This is mainly because too low $N_b$ results in coarse attribute quantization. Consequently, the variation of the generated WCM fails to accurately capture texture deformations.




\end{document}